\documentclass[sensors,article,accept,moreauthors,pdftex]{mdpi} 
\pdfoutput=1

\firstpage{1} 
\pubvolume{19}
\issuenum{6}
\articlenumber{1400}
\pubyear{2019}
\copyrightyear{2019}
\history{Received: 7 February 2019; Accepted: 13 March 2019; Published: 21 March 2019}
\updates{yes} 

\usepackage{geometry}
\usepackage{graphicx}
\usepackage[labelformat=simple]{subfig}

\usepackage{amsmath,amssymb}
\usepackage{listings}
\usepackage{booktabs}
\usepackage{csquotes}
\usepackage{flushend}
\usepackage{booktabs}
\usepackage{multirow}
\usepackage{lineno}
\modulolinenumbers[5]
\usepackage{multicol}
\usepackage{pdflscape}
\usepackage[vlined,algoruled,commentsnumbered]{algorithm2e}

\setitemize{parsep=6pt,itemsep=0pt,leftmargin=*,labelsep=5.5mm}
\setenumerate{parsep=6pt,itemsep=0pt,leftmargin=*,labelsep=5.5mm}
\setlist[description]{itemsep=0mm}

\let\oldnl\nl
\newcommand{\nonl}{\renewcommand{\nl}{\let\nl\oldnl}}

\newlength\lenKwIn
\newcommand\myKwIn[1]{%
  \settowidth\lenKwIn{\KwIn{}}%
  \setlength\hangindent{\lenKwIn}%
  \nonl\hspace*{\lenKwIn}#1\\}

\newlength\lenKwOut

\newcommand{\cc}[1]{\multicolumn{1}{c}{#1}}
\newcommand{\greedy}{{\tt\small greedy}}
\newcommand{\tsp}{{\tt\small FHP}}
\newcommand{\ga}{{\tt\small EA}}
\newcommand{\umari}{{\tt\small UMARI}}

\newcommand{\emp}{{\em empty}}
\newcommand{\potholes}{{\em potholes}}
\newcommand{\jari}{{\em jari-huge}}

\newcommand{\cplus}{{\bf+}}
\newcommand{\cminus}{{\boldmath$-$}}

\Title{An Integrated Approach to Goal Selection in Mobile Robot Exploration}

\Author{Miroslav {Kulich} 
\orcidA{}*, Ji\v{r}{\'\i} Kubal{\'\i}k \orcidB{} and Libor P\v{r}eu\v{c}il}

\AuthorNames{Miroslav Kulich, Ji\v{r}{\'\i} Kubal{\'\i}k and Libor P\v{r}eu\v{c}il}

\address[1]{Czech Institute of Informatics, Robotics, and Cybernetics, Czech Technical University in Prague, 160 00 Prague, Czech~Republic; kubalik@cvut.cz (J.K.); preucil@cvut.cz (L.P.)} 
\corres{\hangafter=1 \hangindent=1.05em \hspace{-0.82em} Correspondence: kulich@cvut.cz}

\abstract{This paper deals with the problem of autonomous navigation of a mobile robot in an unknown 2D environment to fully explore the environment as efficiently as possible. 
We assume a terrestrial mobile robot equipped with a ranging sensor with a limited range and $360^\circ$ field of view.
The key part of the exploration process is formulated as the d-Watchman Route Problem which consists of two coupled tasks---candidate goals generation and finding an optimal path through a subset of goals---which are solved in each exploration step.
The latter has been defined as a constrained variant of the Generalized Traveling Salesman Problem and solved using an evolutionary algorithm. 
An evolutionary algorithm that uses an indirect representation and the nearest neighbor based constructive procedure was proposed to solve this problem.
Individuals evolved in this evolutionary algorithm do not directly code the solutions to the problem.
Instead, they represent sequences of instructions to construct a feasible solution. 
The problems with efficiently generating feasible solutions typically arising when applying traditional evolutionary algorithms to constrained optimization problems are eliminated this way.
The proposed exploration framework was evaluated in a simulated environment on three maps and the time needed to explore the whole environment was compared to state-of-the-art exploration methods.
Experimental results show that our method outperforms the compared ones in environments with a low density of obstacles by up to $12.5\%$, while it is slightly worse in office-like environments by $4.5\%$ at maximum.
The framework has also been deployed on a real robot to demonstrate the applicability of the proposed solution with real~hardware.}

\keyword{path planning; routing; autonomous navigation; generalized traveling salesman problem; evolutionary algorithm}

\begin{document}


\section{Introduction}
\label{sec:intro}

Autonomous mobile robots solve complex tasks in many application areas nowadays.
One~of the most challenging scenarios is search and rescue missions in disaster areas involving large urban areas after an earthquake, a terrorist attack, or a burning house~\cite{ROB21615,Nedjati2016,Bagosi2016}.
Other examples include humanitarian demining, Antarctic, underwater, and space exploration~\cite{Trevelyan2016,montes2015,Murphy2016,Leonard2016}, or navigation in crowded environments~\cite{Choi2014,KimCO18}.
The map of the working environment  in the majority of these scenarios is not known in advance, or the environment dynamically changes significantly so that a priori knowledge is not useful.
For the robot to behave effectively, it has to build a model of the environment from scratch during the mission, determine its next goal, and navigate to it.
This iterative process is called {\em exploration}, and it is terminated whenever the complete map is built.
A natural condition is to optimize the effort needed to perform the exploration, e.g., to minimize the exploration time or the length of the traveled trajectory.

A determination of a next robot goal in each exploration iteration (one exploration step) is called an exploration strategy, which typically consists of three steps. A set of perspective goal candidates is generated first, followed by evaluation of the candidates based on the actual robot position, the current knowledge of the environment, and a~selected optimization criterion (e.g., cost).
The candidate with the best cost is selected as the next goal finally.

A realization of the exploration strategy is the key part of the exploration process as it influences exploration quality significantly---an inappropriate determination of the next goal to be visited may lead to revisiting of already explored places which increases the time needed to finish exploration. 
Therefore, the design of an efficient strategy that determines goals aiming to perform exploration with minimal effort plays an important role.

The exploration problem, and particularly exploration strategies, have been intensively studied in the last twenty years.
The~frontier-based strategy introduced in a seminal work of Yamauchi~\cite{Yamauchi97} considers all points on the {\em frontier}, which is defined as a boundary between a~free and an unexplored space and navigates the robot to the nearest one.
This approach has become very popular as it is simple to implement and it produces reasonable trajectories~\cite{koenig01-greedy,koenig03}. 
Several improvements were suggested.
Holz et al.~\cite{Yamauchi97} segment an already known map, detect rooms in office-like environments and reduce the number of multiple visits of a room by a full exploration of the room once it is entered.
Direction-based selection is presented in~\cite{Mei06energy-efficientmobile}, where the leftmost candidate with respect to the robot position and orientation which is within the current sensing region is selected.
If no such candidate exists, the algorithm picks the closest~one.

While the approaches mentioned above use a single objective to evaluate candidates, another combine multiple criteria.
For example, Gonzalez-Banos and Latombe in~\cite{banos02} mix an effort needed to reach a candidate within an expected area potentially visible from it.
Similarly,  expected information gain expressed as a~change of entropy after performing the action is weighted with the distance to the candidate (distance cost) in~\cite{stachniss05robotics,Perea2017}, while information gain measured according to the expected posterior map uncertainty is used in~\cite{Amigoni10}.
Makarenko et al.~\cite{makarenko02} furthermore introduce the localization utility, which gives preference to places where the robot position can be accurately determined.
Specification of a mixture function and its parameters is a crucial bottleneck of these approaches, and~these are typically set up ad hoc.
Basilico and Amigoni~\cite{Basilico09, basilico2011exploration} therefore introduce a multi-criteria decision-making framework reflecting dependency on particular criteria.
A technique that learns an observation model of the world by finding paths with high information content together with several weight functions evaluating goal candidates is introduced in~\cite{Girdhar2016}. 
An exploration framework based on the use of multiple Rapidly-exploring Random Trees has been introduced recently~\cite{Umari2017}. 
The~authors define a revenue of a goal as a weighted sum of the information gained from exploring the goal and the navigation cost. Moreover, a hysteresis gain is added to prefer goals in robot's vicinity.
A similar approach is proposed in~\cite{Gao2018AnIF}, where a steering angle to the goal is integrated into the utility.

Contrary to the strategies mentioned above, which greedily select the candidate with best immediate cost, Tovar et al.~\cite{Tovar2006} describe an approach where several exploration steps ahead are also considered.
The algorithm goes through a tree structure representing all possible paths the robot may pursue in the given number of steps and searches for the best one employing the branch and bound algorithm.
Although the experimental methodology is not clear, the results show that a greedy approach outperforms the proposed one.
Moreover, the~search state space is large and therefore there is a trade-off between the quality of the solution found and the time complexity and choice has to be made depending on the defined pruning~depth.

An~interesting research stream is a utilization of reinforcement learning.
Zhu et al.~\cite{Zhu2018} introduce Reinforcement Learning supervised Bayesian Optimization based on deep neural networks.
Similarly,~Chen et al.~\cite{chen2018learning} propose a learning-based approach and investigate different policy architectures, reward functions, and training paradigms. 

A~more sophisticated approach to next goal determination is presented in our previous paper~\cite{Kulich2011}.
It is based on the observation that the robot should pass or go nearby all the goal candidates and defines the goal selection problem as the Traveling Salesman Problem (TSP).
That is, the cost of a~candidate $q$ is a~minimal length of the path starting at the current robot position, continuing to the candidate $q$ at first and then to all other candidates.
It was shown that the introduced cost could reduce the exploration time significantly and leads to more feasible trajectories.
The key part of this approach lies in the generation of goal candidates guaranteeing that all frontiers will be explored after visiting all the goal candidates.
An ad hoc procedure is employed, which clusters frontier points by the k-means algorithm and generates candidates as centers of the clusters found.
As k-means considers mutual distances of candidates only and does not take their visibility into account, it can rarely happen that frontiers are not fully covered due to occlusions ({This does not influence completeness of the algorithm as it finishes when no unexplored area remains, and uncovered frontiers will be covered in the next exploration steps when occlusions disappear. 
On the other hand, the quality of the found solution can be degraded.}). 
Moreover, the way how goal candidates are determined has no theoretical relation to the aim of exploration, i.e., traversing all the candidates does not lead to the shortest possible path that explores all frontiers.
A similar approach was then \mbox{used by O{\ss}wald et al.~\cite{Osswald2016}, }who run a TSP solver on a priori user-defined topological map. 
The~authors, in consensus with our results, experimentally demonstrated that this method significantly reduces the exploration time.

Faigl and Kulich~\cite{Faigl2013} formulate candidates' generation as a variant of the Art Gallery Problem with limited visibility, which aims to find a minimal number of locations covering all frontiers. The~proposed iterative deterministic procedure called Complete Coverage follows the idea of a generation of samples covering free curves proposed by Gonzalez-Banos and Latombe~\cite{banos02} and guarantees full coverage of frontiers.
Nevertheless, candidates generation and goal selection are still independent processes, i.e., candidates are not generated with respect to the cost of a path visiting all the candidates.

To the best of our knowledge, the only attempt to join these two processes into a single procedure is presented in Faigl~et~al.~\cite{Faigl2014}, where the goal selection task is formulated as the Traveling Salesmen Problem with Neighborhoods and a two-layered competitive neural network with a variable size to solve the problem is proposed.
The presented results show that the approach is valid and it provides good results for open-space environments and longer visibility ranges and for office-like environments and small visibility ranges.
On the other hand, the approach is very computationally demanding, which limits its deployment in real applications.

The research presented in this paper continues in the direction outlined in our previous works as it introduces an integrated solution to goal candidates' generation and goal selection.
Novelty and contribution of the paper stand mainly in the following:
\begin{itemize}[leftmargin=*,labelsep=5.8mm]
\item We formulate the objective of the integrated approach to candidates generation and goal selection as the d-Watchman Route Problem, which enables a theoretically sound interpretation of the integrated approach to the goal determination problem.
Our solution to the problem then leads to a definition of the objective of a goal selection itself as a variant of the Generalized Traveling Salesman Problem (GTSP).
\item The introduced GTSP variant involves additional constraints to the original GTSP, which renders standard GTSP solvers inapplicable here.   
A novel evolutionary algorithm taking into account the added constraints is introduced. It uses an indirect representation and an extended nearest neighbor constructive procedure which circumvent the candidate solutions feasibility issue encountered when using traditional evolutionary algorithms with direct encodings.
\item A novel approach for generating nodes/vertices for GTSP is presented as a mixture of the techniques for goal candidates generation mentioned above.
\item The whole exploration framework is evaluated in a simulated environment and compared to the state-of-the-art methods. Moreover, we implemented the framework on a real robot to show that the proposed strategy is applicable in real conditions.

\end{itemize}

The fundamental single-robot exploration in a 2D environment, which is the main interest of the paper, has several extensions.
One of these is the multi-robot case in which coordination of multiple robots is studied. 
Several goal selection strategies for multi-robot coordination based on different principles were introduced by many authors. 
A Markov process to model exploration using the transition probabilities to consider environment characteristics is proposed in~\cite{andre13iros}, while~a greedy approach is introduced in~\cite{Burgard2005} and bio-inspired goal selection using a hybrid pheromone and anti-pheromone signaling mechanism is presented in~\cite{Ravankar2016}.
A goal selection formulated as a multi-vehicle variant of the Travelling Salesman Problem is presented in~\cite{Faigl2012}, while several exploration strategies are compared in~\cite{Faigl2013}.

The previously mentioned exploration approaches assume that robot position is known, either provided by GPS or some SLAM (Simultaneous Localization and Mapping) algorithm.
One of the first attempts to deal with localization uncertainty is presented in~\cite{Yamauchi1998}. 
After that, several approaches employing  Bayes filter were proposed~\cite{lidoris2011state,Tovar2006,papachristos2017uncertainty}.
With the increasing popularity of aerial robots and achievements in 3D sensing, exploration in three dimensions has been also studied (see~\cite{Selin2019,papachristos2017uncertainty,delmerico2017active,Bircher2018}).

The rest of the paper is organized as follows.
The problem definition is presented in Section~\ref{sec:problem}. 
The proposed exploration framework is described in Section~\ref{sec:framework}.
The evolutionary algorithm designed for the constrained variant of the Generalized Traveling Salesman Problem and employed for goal determination is introduced in Section~\ref{sec:ea} and the experimental evaluation of the method is presented in Section~\ref{sec:experiments}.
Finally, concluding remarks and future directions are summarized in~\ref{sec:conclusion}.


\section{Problem Definition}
\label{sec:problem}

Assume a fully localized autonomous mobile robot equipped with a ranging sensor with a fixed, limited range (e.g., a laser range-finder) and $360^\circ$ field of view operating in an unknown flat environment. 
Exploration is defined as the process in which the robot is navigated with the aim to build a map of the surrounding space to collect information about this space.
The map is built incrementally as sensor measurements are gathered and it serves as a model of the environment for further exploration steps.

\begin{figure}[H]
\begin{center}
\includegraphics[width=0.35\columnwidth]{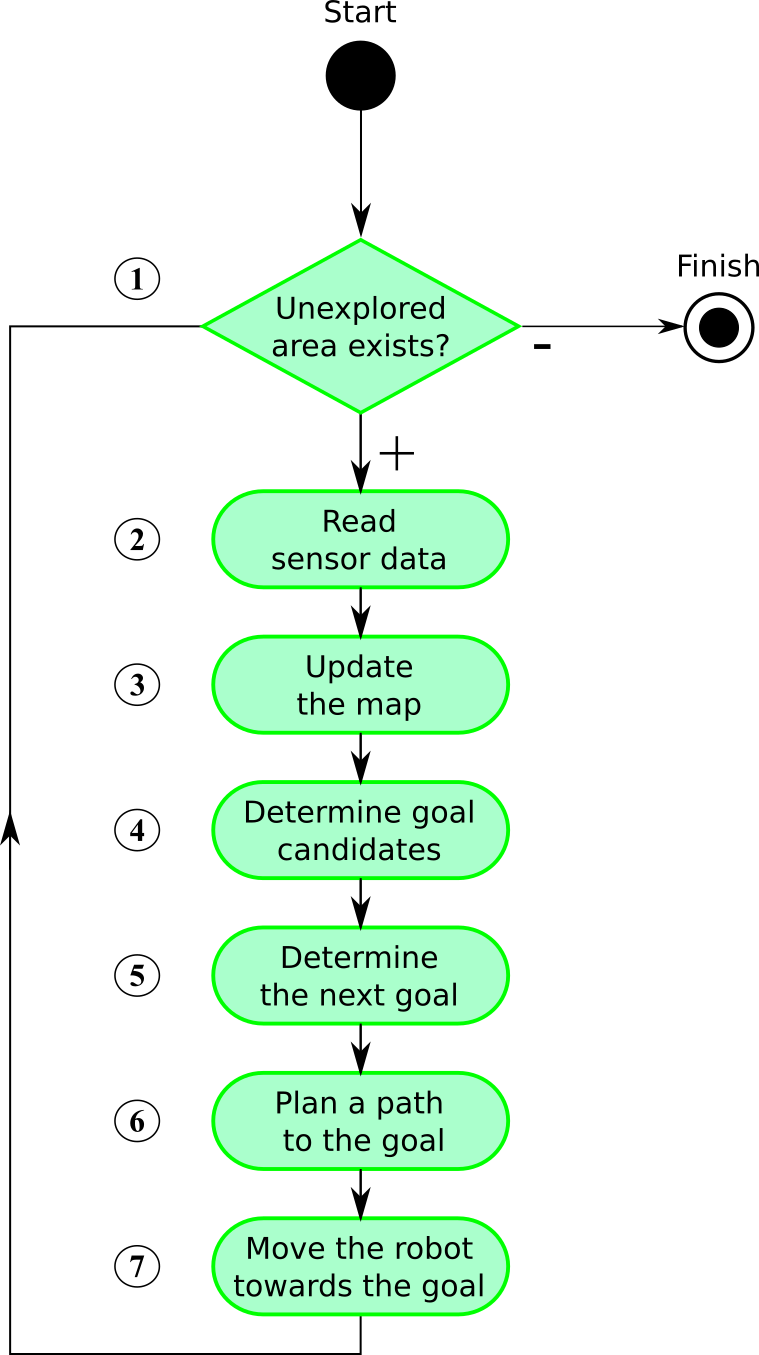}
\end{center}
\caption{The exploration process.}
\label{sc:exploration}
\end{figure}

The whole exploration process is summarized in Figure~\ref{sc:exploration}.
The algorithm consists of several steps that are repeated until no unexplored area remains. We assume that the environment is bounded. 
The~exploration process is thus finished when there is no remaining unexplored area accessible to the robot (step~1). 
Accessibility is an essential condition as interiors of obstacles are inaccessible and thus remain unexplored. 

The process starts with reading actual sensory information (step~2).
The map is updated after some data processing and noise filtering (step~3).
New goal candidates are determined afterward (step~4) and a next goal for the robot is assigned using a defined cost function (step~5).
The shortest path from the robot's current position to the goal is found next (step~6) and  the robot is navigated along the path (step~7).

While steps 2,3,6, and 7 belong to fundamental robotic tasks and many solutions to these already exist, the key part heavily influencing efficiency of the exploration process is the \textbf{next goal assignment problem} (step~5), formally defined as follows: 

Given a current map $\boldsymbol{M} \subset {\cal R}^2$, a robot position $p\in \boldsymbol{M}$, and  $n$ goal candidates located at positions $\boldsymbol{G}=\{g_1,\ldots, g_n\} \subset \boldsymbol{M}$.
The problem is to determine a goal $g\in\boldsymbol{G}$ that minimizes the total required time (or the traveled distance) needed to {explore the whole environment}. 

Goal assignment together with the determination of goal candidates in step~4 is called \textbf{exploration strategy}. 
More formally, exploration strategy aims to find a policy $\pi: \boldsymbol{M} \times \cal{M} \rightarrow \boldsymbol{M}$, where $\cal{M}$ is a set of all possible maps and $\boldsymbol{M} \in \cal{M}$ is a map. 
Given the robot position $p\in \boldsymbol{M}$ and the map $\boldsymbol{M}\in \cal{M}$, the~policy determines the next goal $g^* \in \boldsymbol{M}$ so that the following cost is minimized:
\begin{equation}
\pi = \arg\min_a \sum_{t=0}^{T^a-1}cost(p^a_t, p^a_{t+1}).
\end{equation}

$T^a$ is the time needed to explore the whole environment and $p^a_{t+1}$  is robot position at time $t+1$ when the policy $a$ is followed.
$cost(x,y)$ is a cost of movement from $x$ to $y$.
Notice that, if $cost(x,y)$ is set to $1~~\forall x,y$, then time is minimized, while setting $cost(x,y)$ as the distance of a path from $x$ to $y$ leads to optimization of a total travelled distance.
Similarly to the exploration literature, we assume that a robot moves with a constant velocity and thus traveled distance linearly depends on a total traveling time so these two can be interchanged.

Determination of the optimal policy is not possible in general because $cost$ and $T^a$ cannot be exactly computed without knowledge of a structure of the unexplored environment. 
The aim is thus to find a strategy that leads to an exploration of the whole environment in the shortest possible time.   

\begin{figure}[H]
    \centering
    \includegraphics[width=0.4\columnwidth]{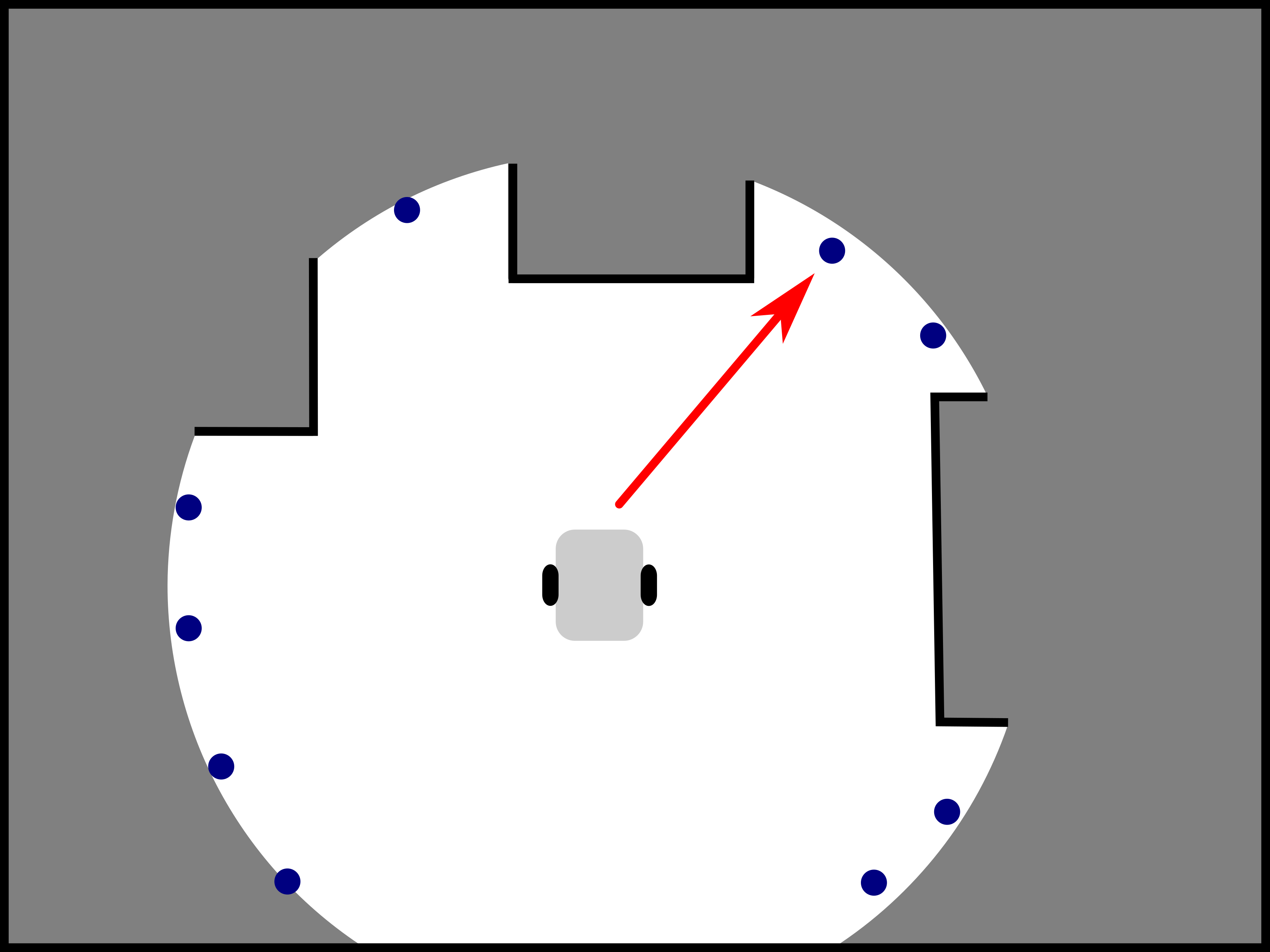}
    \caption{Illustration of the exploration\ strategy objective.} 
    \label{fig:goal_selection}
    \end{figure}

The objective of the exploration strategy is illustrated in~Figure~\ref{fig:goal_selection}.  
Goal candidates (blue points) are generated on the border of an already explored area (white) and an unexplored space (grey) or in its vicinity. 
The goal assignment then selects the candidate minimizing some predefined penalty function as the next goal to which the robot is navigated (the red arrow points to the chosen goal).
A~typical penalty function is, for example, {\em distance}, i.e., the candidate nearest to the robot is selected.

Realization of the particular steps of the exploration problem, especially the proposed exploration strategy, is detailed in Section~\ref{sec:framework}.

Note that it is inefficient to invoke goal determination after the whole path is traversed and the current goal is reached.
The experimental evaluation made by Amigoni~et~al.~\cite{Amigoni2013} indicates that it is better to run this decision-making process continuously, at some fixed frequency.
The~authors show that increasing decision frequency generally increases performance, but when the frequency is too high, performance degrades due to increased computational effort.
Moreover, higher frequencies lead to oscillations of a robot's movement causing longer trajectories and exploration time.
Best~results in their paper were achieved with the frequency of 0.6~Hz, while the authors of~\cite{NewmanBosseLeonard03} \mbox{use the frequency of 0.25~Hz.}

The exploration algorithm can be implemented in two threads: the first one communicating with the robot hardware and containing sensor processing, map building, and robot control with high frequency of units or tens of Hz and the decision-making process incorporating goal candidates generation, goal selection and path planning in the second one, which is triggered with a lower frequency.

\section{Exploration Framework}
\label{sec:framework}

\begin{figure}[H]
\centering
\includegraphics[width=0.6\columnwidth]{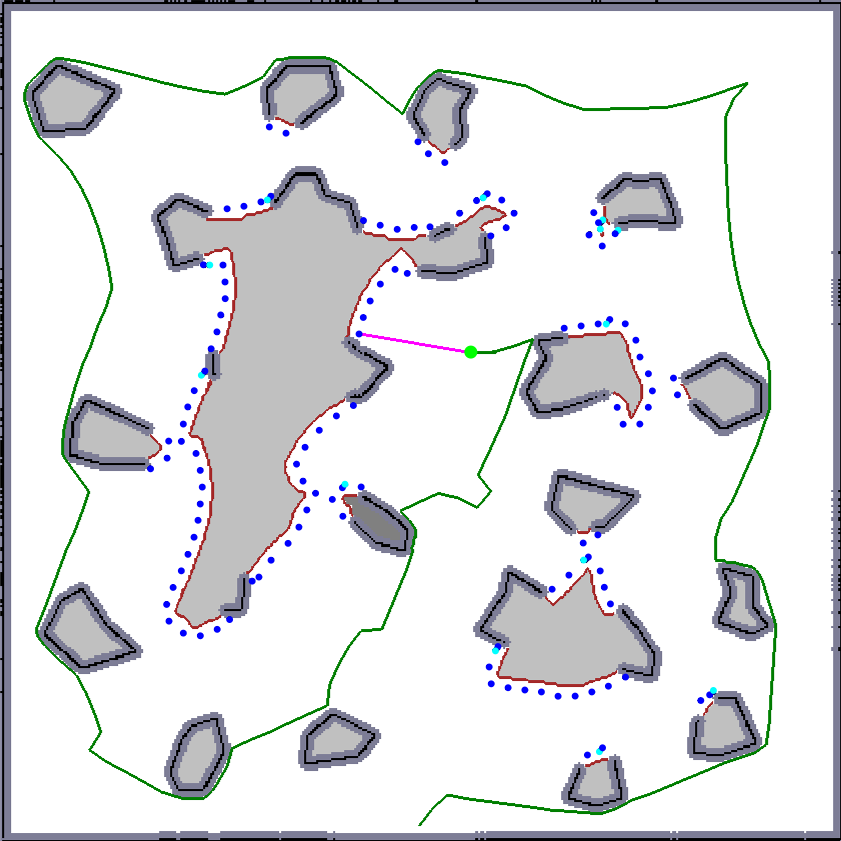}
\caption{An exploration step: the robot (the green circle) selects one of the goal candidates (the blue dots; note that two different hues are used to visually distinguish overlapping dots.) as a next goal and plans a path to it (the pink curve). The green curve represents the already traversed trajectory.}
\label{fig:terms}
\end{figure}

The proposed exploration framework described in Algorithm~\ref{sc:framework} is derived from Yamauchi's frontier based approach~\cite{Yamauchi97}.
The approach employs an occupancy grid~\cite{elfes89} for map representation, which divides the working space into small rectangular cells.
Each cell stores information about the corresponding piece of the environment in the form of a probabilistic estimate of its state, see Figure~\ref{fig:terms}. 
Assuming that the map is static and individual cells are independent, a cell can be updated using a Bayes rule as described in~\cite{elfes89}:
$$ Bel(m_t^{[xy]}) = \eta p(z_t|m_t^{[xy]}) Bel(m_{t-1}^{[xy]}),$$
 where $\eta$ is a normalization constant ensuring that probabilities of all possible states of $m_t^{[xy]}$ sum to 1, $p(z_t|m_t^{[xy]})$ is {\em a sensor model}, and $Bel(m_{t-1}^{[xy]})$ is the current believe in the state of $m^{[xy]}$ determined in the previous step. 
 No a priori information about the environment is provided, therefore, $Bel(m_{0}^{[xy]})$ is set to $\frac{1}{2}$ for all cells.

\begin{figure}[H]
  \centering
  \includegraphics[width=0.3\columnwidth]{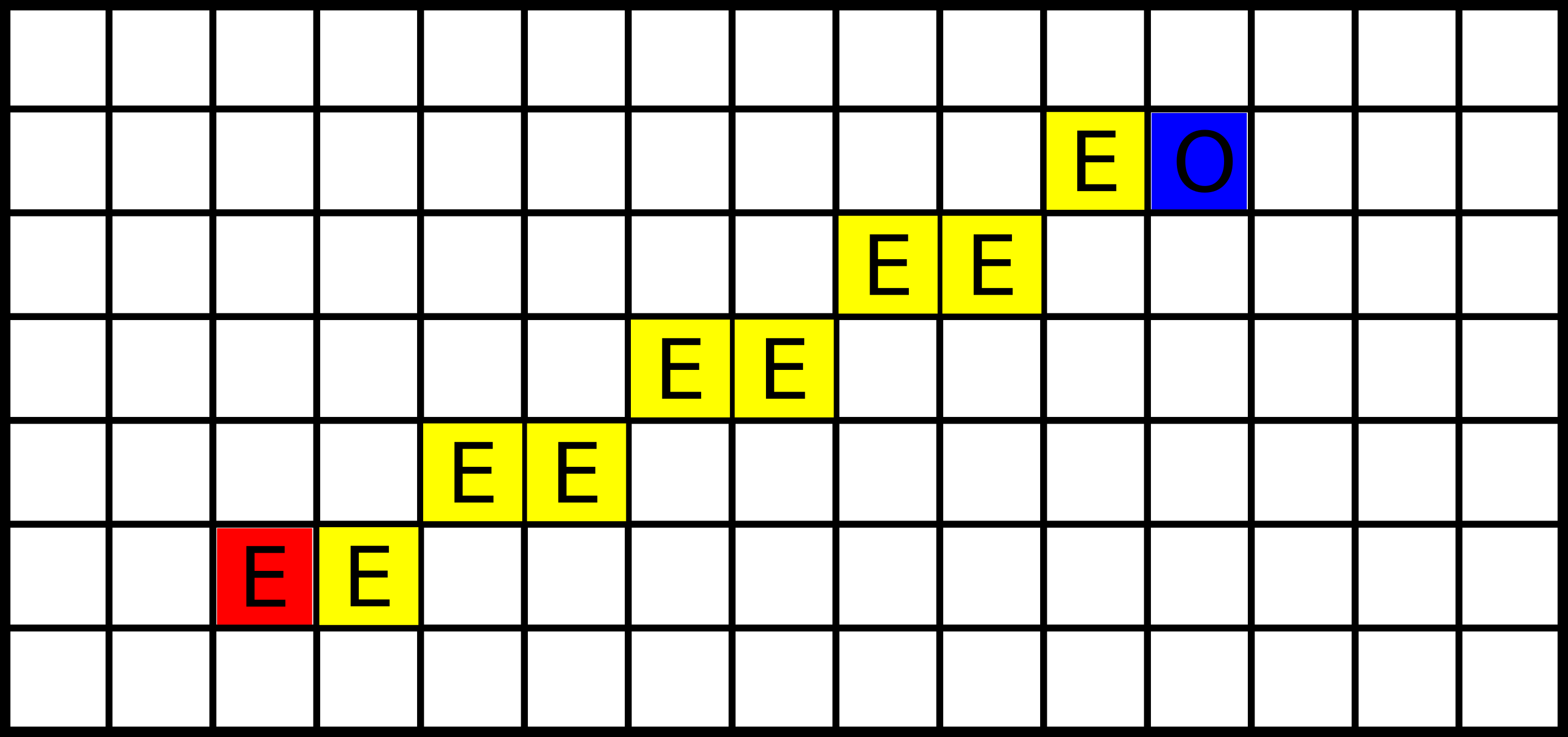}
  \caption{Sensor model. The robot (the red cell) measures an obstacle at the blue cell. The numbers in the cells represent sensor model values.}
  \label{fig:sensor_model}
  \end{figure}

A precision of contemporary ranging sensors is in order of centimeters, which is lower than or equal to the size of a grid cell. 
Therefore, a simple sensor model is used: $p(z_t|C)=O$ for the grid cell $\cal C$ corresponding to the sensor measurement, while $p(z_t|m^{[xy]})=E$ for cells $m^{[xy]}$ lying on an abscissa between the current robot position $
\cal S$ and $\cal C$, see Figure~\ref{fig:sensor_model}. 
For a simulated or ideal sensor, we set $O=1$, and $E=0$. 
For real sensors, where a fusion of measurements during time suppresses an influence of sensor noise, the values are set closer to $\frac{1}{2}$, e.g., $O=0.3$, and $E=0.7$. 

Occupancy grid cells are consequently segmented into three categories by application of two thresholds on their probability values: free (represented by a white color in Figure~\ref{fig:terms}), occupied (black), and unexplored (light gray). 
The occupied cells are inflated by Minkowski sum (the dark gray areas in Figure~\ref{fig:terms}) enabling to plan non-colliding paths for a circular-shaped robot (the robot does not collide with obstacles if its center lies outside inflated areas).
The frontier based approach detects {\em frontier cells} (represented by a brown color in Figure~\ref{fig:terms}), i.e., the reachable free grid cells that are adjacent with at least one cell that has not been explored yet (line~\ref{es2}).
The {\em frontier} is a~continuous set of frontier cells such that each frontier cell is a~member of exactly one frontier.

Given determined frontiers, the exploration strategy determines a next goal the robot is navigated to as detailed in Section~\ref{sec:strategy}.
The shortest path to this goal is determined by application of some standard planning algorithm like Dijkstra’, A*, Voronoi Diagram or wave-front propagation techniques like Distance Transform~\cite{Faigl2015}.
Finally, navigation of the robot to the goal is realized by a local collision avoidance algorithm, e.g., Smooth Nearness-Diagram Navigation~\cite{durham08}. 

\subsection{Proposed Exploration Strategy}
\label{sec:strategy}
While Yamauchi~\cite{Yamauchi97} greedily selects the frontier cell nearest to the current robot position, the proposed approach is more sophisticated as depicted in Algorithm~\ref{sc:framework}.
The idea behind it is motivated by the fact that it is not necessary to visit a frontier to explore a new area surrounding it.
Instead, visiting places from which all frontier cells are visible is sufficient. In fact, seeing slightly \enquote{behind} frontiers is requested to get new information. We ensure this by generating possible goals closer to frontiers than at visibility range as will be described below.

The realization of this idea leads to the formulation of next goal determination as a variant of the Watchman Route Problem~\cite{Mitchell2013}:
Given a connected domain $P$, the Watchman Route Problem is to compute a shortest path or tour for a mobile guard that is required to see every point of $P$.
In our case, seeing only all frontier cells (i.e., boundary of a domain containing all free cells) assuming limited visibility is requested only.
Nevertheless, the problem, called as d-Watchman Route Problem, remains NP-hard~\cite{Chin1988} and therefore its solution is split into a solution of two sub-problems:
\begin{itemize}[leftmargin=*,labelsep=5.8mm]
\item \textbf{Generation}: of a set of goal candidates so that union of their visibility regions (i.e., areas visible from the particular goal candidates) contains all frontier cells, and
\item \textbf{Construction} of a route connecting  a subset of the goal candidates, whose visibility regions cover all frontier cells and a path traversing all of them from a current robot position is minimal.
\end{itemize}

An example of a set of goal candidates is given in Figure~\ref{fig:terms}, where the candidates are represented as blue points.
The purpose of the Generation phase is to construct a huge number of goal candidates which is then used as an input to the Construction phase and which guarantees that a feasible solution is constructed.
This huge number of candidates, which is an order of magnitude higher than in~\cite{Kulich2011}, gives big flexibility to the Construction algorithm to produce good solutions of the superior d-Watchman Route Problem as it can choose from many possible candidates.
Given an optimal solver in the Construction phase, a solution found by it converges with an increasing number of candidates to an optimal solution of the d-Watchman Route Problem.

\subsubsection{Goal Candidates Generation}
\label{sec:generation}
A na\"{\i}ve approach to the Generation problem  takes all frontier cells as goal candidates.
This leads to hundreds or thousands of goal candidates and high computational burden of determination of a path connecting them.
Instead, the proposed algorithm generates an order of magnitude lower number of candidates.
It starts with segmentation of an occupancy grid (Algorithm~\ref{sc:framework}, line~\ref{es1}) and detection of all frontier cells (line~\ref{es2}).
A connected string of frontier cells of each particular frontier is created then (line~\ref{es3}).
This can be done by one of boundary tracking algorithms known from image processing for extracting boundaries for images.
Namely, Moore-neighbor tracing algorithm~\cite{reddy2012evaluation} is employed in \mbox{our case.}

The ordering of frontier cells helps to efficiently determine goal candidates (lines~\ref{es4}--\ref{es15}), which is done for each frontier contour $\boldsymbol{F}$  in two stages.

In the first stage, candidates are generated uniformly.
Note that the shape of a frontier is not smooth in general due to the discretization of the world into grid cells and sensory data noise.
This~often causes occlusions of frontier cells lying in a visibility range.
Therefore, candidates are not generated on the frontier, but at the distance $d$ from $\boldsymbol{F}$, which is a fraction of a visibility range:
$\boldsymbol {F}$ is inflated by a disk with a radius $d$ using Minkowski sum (line~\ref{esA1}),
a contour of the inflated frontier is detected by the boundary tracking algorithm (line~\ref{esA2}) and every $k$-th frontier cell of the contour is taken as a goal candidate (line~\ref{esA13}).
Coverage, i.e., frontier cells visible from the candidate, is computed (lines~\ref{esA14}--\ref{esA16}) and subtracted from the set of uncovered frontier cells (line~\ref{esA17}), which is initially set to $\boldsymbol{F}$ (line~\ref{es7}).

It can happen that a frontier is, due to occlusions, not fully covered by visibility regions of candidates generated uniformly.
Therefore, a dual sampling scheme is utilized to cover uncovered frontier cells:
a random not yet covered cell is selected from the frontier $\boldsymbol {F}$ (line~\ref{es9}), a set of free cells visible from it is determined (line~\ref{es10}), and a goal candidate is generated randomly from these free cells (lines~\ref{es11} and \ref{es12}).
Finally, the set of uncovered frontier cells in updated in the same way as in the first stage (line~\ref{es15}).
The dual sampling is repeated until no uncovered frontier cell remains (line~\ref{es8}).
Candidates generated from a single frontier $\boldsymbol{F}$ form {\em a cluster} $\boldsymbol{G_F}$.

\vspace{12pt}
\LinesNumbered
\SetKwFor{Times}{}{times do}{end}
\DontPrintSemicolon
\begin{algorithm}[H]
\small
\KwIn{$\boldsymbol{M}$ -- occupancy grid}
\myKwIn{$\delta$\hspace{1em}-- visibility range}
\myKwIn{$r$\hspace{1em}-- current robot position}
\KwOut{$g^*$ -- next goal}
\vspace{-0.5em}
\nonl\hrulefill\\
$\left<\boldsymbol{M}_{free},\boldsymbol{M}_{unknown},\boldsymbol{M}_{empty}\right> \leftarrow$ segmented $\boldsymbol{M}$\;\nllabel{es1}
$\boldsymbol{E} \leftarrow$ frontier cells of $\boldsymbol{M}$\;\nllabel{es2}
${\cal F} \leftarrow $ contours of $\boldsymbol{E}$\;\nllabel{es3}
\ForEach{$\boldsymbol{F}\in {\cal F}$}{\nllabel{es4}
  $\boldsymbol{G_F} \leftarrow \emptyset$ \nllabel{es6}\tcp*{goal generated by $\boldsymbol{F}$}
  $\boldsymbol{U} \leftarrow \boldsymbol{F}$ \nllabel{es7} \tcp*{uncovered frontier cells of $\boldsymbol{F}$}
  $\boldsymbol{I} \leftarrow$ inflate $\boldsymbol{F}$\;\nllabel{esA1}
  $\boldsymbol{T} \leftarrow $ contours of $\boldsymbol{I}$\;\nllabel{esA2}
  \ForEach{$k$-th $g\in\boldsymbol{T}$}{\nllabel{esA3}
    $\boldsymbol{G_F} \leftarrow \boldsymbol{G_F} \cup \{g\}$\;\nllabel{esA13}
    $\boldsymbol{V} \leftarrow \boldsymbol{M}_{free} \cap \boldsymbol{R}(g)$, where \;\nllabel{esA14}
    \nonl\hspace{1em}$\boldsymbol{R}(g)$ is a set of cells visible from $g$\;\nllabel{esA15}
    $g_{coverage} \leftarrow \boldsymbol{F} \cap \boldsymbol{R}(g)$\;\nllabel{esA16}
    $\boldsymbol{U} \leftarrow \boldsymbol{U} \setminus g_{coverage}$\;\nllabel{esA17}
  }

  \While{$\boldsymbol{U} \neq \emptyset$} {\nllabel{es8}
    $s\leftarrow$ random cell from $\boldsymbol{U}$\;\nllabel{es9}
    $\boldsymbol{V} \leftarrow \boldsymbol{M}_{free} \cap \boldsymbol{R}(s)$, where \;\nllabel{es10}
    \nonl\hspace{1em}$\boldsymbol{R}(s)$ is a set of cells visible from $s$\;\nllabel{es11}
    $g\leftarrow$ random cell from $\boldsymbol{V}$\;\nllabel{es12}
    $\boldsymbol{G_F} \leftarrow \boldsymbol{G_F} \cup \{g\}$\;\nllabel{es13}
    $g_{coverage} \leftarrow \boldsymbol{F} \cap \boldsymbol{R}(g)$\;\nllabel{es14}
    $\boldsymbol{U} \leftarrow \boldsymbol{U} \setminus g_{coverage}$\;\nllabel{es15}
  }
  ${\cal G} \leftarrow {\cal G} \cup \boldsymbol{G_F}$\;\nllabel{es16}
}

$D\leftarrow$ distance matrix of goals in ${\cal G}$\;\nllabel{es17}
$tour \leftarrow f({\cal G},D)$\;\nllabel{es18}
$g^*\leftarrow$ second node of $tour$\;\nllabel{es19}
\KwRet{$g^*$}\;\nllabel{es20}
\caption{Proposed exploration strategy}
\label{sc:framework}
\end{algorithm}

\subsubsection{Route Construction}
\label{sec:construction}
The Route construction phase starts with the computation of a distance matrix of all generated goal candidates by running Dijkstra's algorithm~\cite{Cormen2009} on an adjacency graph of the free cells for each candidate (line~\ref{es17}).
Note that utilization of all-pairs shortest path algorithms (e.g., Johnson's~\cite{Johnson1977} with $O(V E \log V)$, where $V$ is a number of vertices and $E$ is a number of edges) is not effective, as~these compute distances among all free cells, while complexity of $K$ runs of Dijkstra's algorithm is $O(K E+ KV\log V)$, where a number of goal candidates $K<<N$.

Finally, the determined set of clusters, each containing a set of corresponding goals with their coverages, and the distance matrix of the goals are used to find the best candidate as the next goal to move to in the next iteration of the exploration process (line~\ref{es18}--\ref{es20}).
The problem of finding such goal involves the solution of an optimization problem of constructing a shortest possible tour starting from the current robot position and leading through a subset of goals subject to the constraint that all frontier cells are covered by the selected goals (line~\ref{es18}). 
The next goal $g^*$ is then determined as the second node ({Note that the first node of the tour is robot's position.}) of $tour$ (line~\ref{es19}), which is returned (line~\ref{es20}).

The optimization problem at line~\ref{es18} can be formulated as a variant of the GTSP where the formulation of the general GTSP is:
Given a set $\boldsymbol{G}$ of nodes partitioned into $m$ non-empty clusters and a matrix $D$ of their mutual distances, the objective is to find a subset $\boldsymbol{S}$ of $\boldsymbol{G}$ so that (a) $\boldsymbol{S}$ contains \emph{exactly one} node from each cluster and (b) a tour connecting all nodes from $\boldsymbol{S}$ is the shortest possible. 
The~GTSP is known to be NP-hard as it reduces to the NP-hard TSP when each cluster consists of exactly one node.

Here, we solve a constrained variant of the GTSP where the final tour contains \emph{at least one node from each cluster} and must satisfy a constraint that all frontiers are fully covered by goals visited within the tour.
To formulate the problem formally, assume
\begin{itemize}[leftmargin=*,labelsep=5.8mm]
\item a set of $m$ frontiers $\mathbf{F}=\{F^i\}_{i=1}^m$, where each frontier $F^i$ is a set of frontier cells $F^i=\{f^i_j\}$,
\item a set of $n$ goal candidates $\mathbf{G}$ clustered into $m$ clusters. Each cluster $G^i=\{g^i_j\}^{n_i}_{j=1}$ associated to a frontier $F^i$ is represented by $n_i$ candidate goals,
\item coverage $\mathbf{R}(g^i_{j}) \subset F^i$ for each $g^i_j$,
\item a current position $r$ of the robot,
\item a matrix $D$ of mutual distances between all goal candidates.
\end{itemize}

The aim is to find a tour $t$ such that
\begin{equation}
t = \arg\min_\mathbf{T}\sum_{k=0}^{p-1} D(g^{i_k}_{j_k},g^{i_{k+1}}_{j_{k+1}}),
\label{eq:gtspc}
\end{equation}
where $\mathbf{T}$ is a set of all tours $\tau =\left<g^{i_0}_{j_0}, g^{i_1}_{j_1}, g^{i_2}_{j_2}, \dots,\right.$ $\left. g^{i_p}_{j_p}\right>$  satisfying 

\begin{eqnarray}
&&g^{i_0}_{j_0}=r \text{~~~~and}\label{eq:cond1}\\
&&\bigcup\limits_{k=0}^{n} \mathbf{R}(g^{i_k}_{j_k}) =  \bigcup\limits_{i=1}^{m} F^i\label{eq:cond2}
\end{eqnarray}

We call this problem Generalized Travelling Salesman Problem with Coverage and denote \mbox{it GTSPC.}

A lot of attention has been paid by researchers to solve the GTSP. There are approaches transforming the GTSP to the TSP and solving the related TSP \cite{dimitrijevic97,ben-arieh03}.
Other approaches formulate the GTSP as an integer linear program (ILP) and employ exact algorithms for solving the ILP \cite{laporte83,fischetti97,kara12}.
Due to the NP-hardness of the GTSP problem, many heuristic, metaheuristic and hybrid approaches have been developed in the past decade as well.
For example, a metaheuristic global search based on imperialist competitive algorithm inspired by a socio-political scheme and hybridized with a local search procedure is introduced in~\cite{Ardalan2015}. 
El Kraki et al.~\cite{ElKrari2017} present a simple heuristic that clusters input cities, finds their barycenters to determine an order of the clusters and determines the best city of each cluster.
\cite{karapetyan12} provides a survey of neighborhoods and local search algorithms for the GTSP.

A~hybrid genetic algorithm with the random-key indirect representation that enforces only feasible solutions created by genetic operators has been proposed in \cite{snyder06}.
A~memetic algorithm with a powerful local search procedure making use of six different local search heuristics was proposed in \cite{gutin10}.
Various versions of Ant Optimization Algorithms were used to solve the GTSP as well \cite{mou11,pintea13,reihaneh12}.

In this work, we assume the GTSPC where the tour can contain one or more cities from each cluster.
This is implied by the constraint imposed on the selected subset of goal candidates that they must completely cover all frontiers.
All published approaches to the GTSP consider the unconstrained case, where the tour contains exactly one city from each cluster. 
These approaches thus do not provide valid solutions for the considered GTSPC and it is not possible to adapt them straightforwardly to this problem.
We, therefore, propose an evolutionary algorithm, described in the following section, \mbox{to efficiently} solve the constrained variant of the GTSP--GTSPC.

\section{Proposed Approach to Solve the GTSPC}
\label{sec:ea}

The exploration strategy aims to determine a goal to which the robot will move next, which leads to finding a shortest possible tour through a properly selected subset of goals as discussed in {Section}~\ref{sec:generation} (see lines~\ref{es18}--\ref{es19} of {Algorithm}~\ref{sc:framework}). 
We defined this optimization problem as GTSPC, i.e., the~next goal is the second node of a tour starting from the current position and minimizing Equation~(\ref{eq:gtspc}) while satisfying Equations~(\ref{eq:cond1}) and~(\ref{eq:cond2}).

The~proposed approach to solve GTSPC is based on the evolutionary algorithm with indirect representation and extended nearest neighbor constructive procedure (IREANN), originally proposed 
for a symmetric TSP~\cite{kubalik14}. IREANN is an evolutionary algorithm particularly suited for solving routing and sequencing problems. When searching for a good solution to the routing problem, it directly exploits the short or low-cost links. Section~\ref{sec:ireann} describes the main idea behind IREANN and basic components of the algorithm on an example of the TSP. 
The proposed adaptation of the algorithm to the GTSPC variant considered in this work is introduced in Section~\ref{sec:ireann-gtspc}.

\subsection{Original IREANN Algorithm}
\label{sec:ireann}
IREANN can be considered an extension of the \emph{nearest neighbor} (NN) constructive algorithm used for solving the routing problems such as the TSP. The~main idea is to explore a larger set of candidate solutions than the NN algorithm while making use of the shortest links as much as possible.

\begin{figure}[H] 
\begin{center}
\subfloat[][]{\includegraphics[width=0.3\columnwidth]{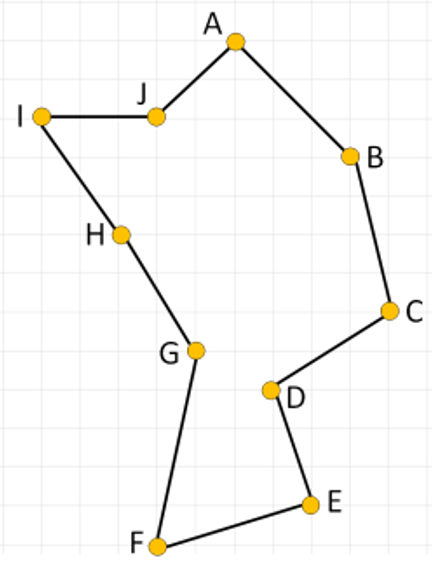}\label{fig:ireann_motivationA}}\hfill
\subfloat[][]{\includegraphics[width=0.3\columnwidth]{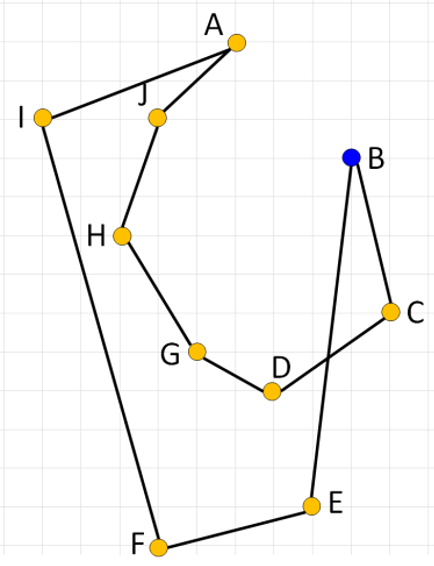}\label{fig:ireann_motivationB}}\hfill
\subfloat[][]{\includegraphics[width=0.3\columnwidth]{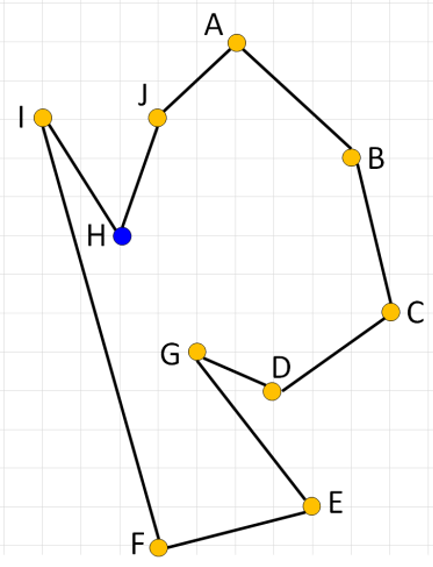}\label{fig:ireann_motivationC}}
\end{center}
\vspace{-1em}
\caption{Illustration of an ineffectiveness of the standard nearest neighbor constructive procedure on a simple instance of the TSP. (\textbf{a}) shows an optimal solution; (\textbf{b}) shows a solution constructed by the nearest neighbor when started from node B; (\textbf{c}) shows a solution constructed when the procedure starts from node~H.}
\label{fig:ireann_motivation}
\end{figure}

The standard NN algorithm starts the tour in a randomly chosen working city, $s$. Then,~ it~repeatedly connects city $s$ with its nearest neighbor $v$ chosen from the set of available nodes at the moment $\mathbf{A}$
\begin{equation}
v \leftarrow \arg\min_{a \in \mathbf{A}} dist(s,a)
\label{eq4}
\end{equation}
and the city $v$ becomes the current working city $s$ for the next iteration. The~process stops when all cities have been visited. We can generate up to $N$ different tours, where $N$ is the number of cities considered, and take the best tour as the final solution. This algorithm is easy to implement, quickly yields a short tour, but it produces suboptimal tours very often due to its greedy nature. An~example of such a case is illustrated in Figure~\ref{fig:ireann_motivation} with an optimal solution (Figure~\ref{fig:ireann_motivation}a) and two suboptimal solutions generated by the NN algorithm started from node B (Figure~\ref{fig:ireann_motivation}b) and H (Figure~\ref{fig:ireann_motivation}c),
respectively. Note~that~the algorithm initialized in either of the ten starting nodes is not able to produce the optimal~solution.

\begin{figure}[H] 
\begin{center}
\subfloat[][]{\includegraphics[width=0.3\columnwidth]{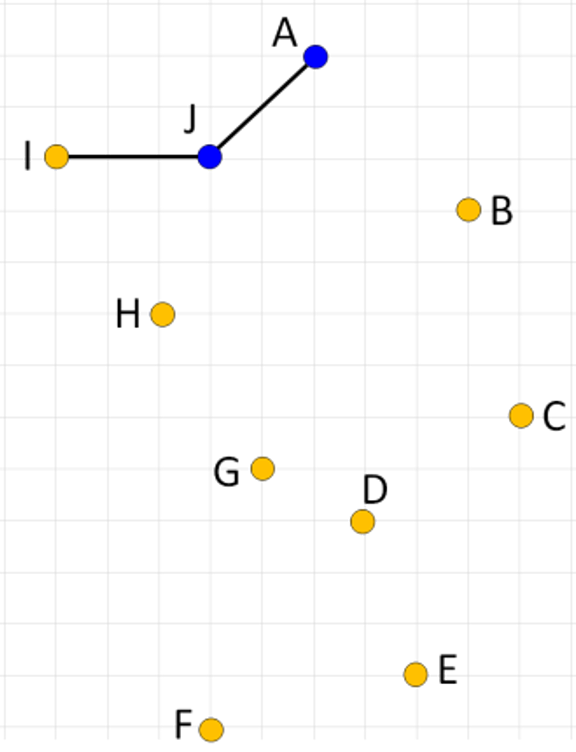}\label{fig:ireann_exampleA}}\hfill
\subfloat[][]{\includegraphics[width=0.3\columnwidth]{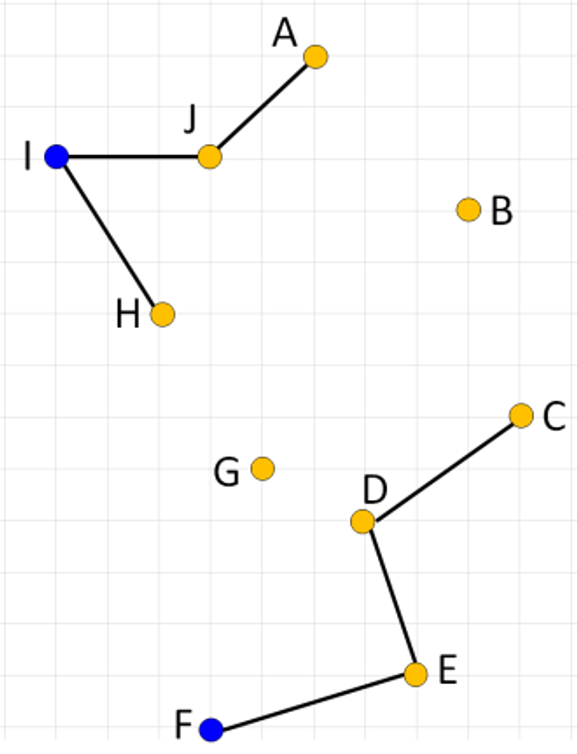}\label{fig:ireann_exampleB}}\hfill
\subfloat[][]{\includegraphics[width=0.3\columnwidth]{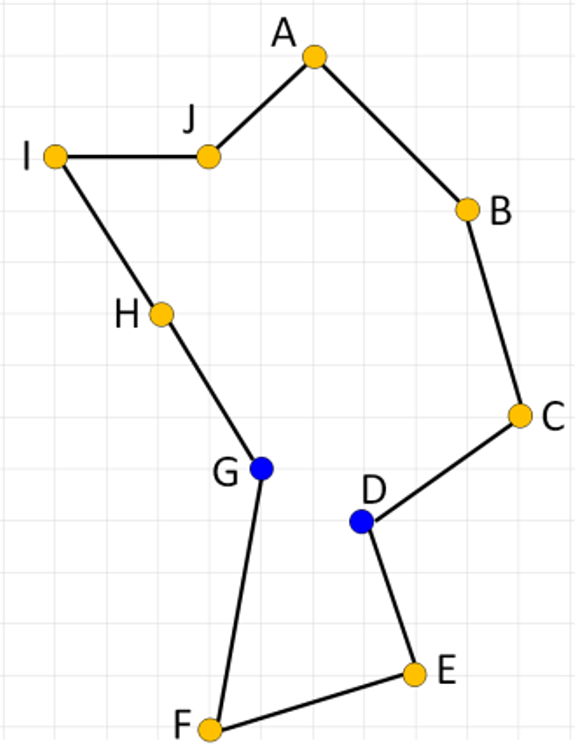}\label{fig:ireann_exampleC}}
\end{center}
\vspace{-1em}
\caption{{Extended nearest} neighbor constructive procedure applied to cities in the following order \texttt{AJCEIFBHGD}. (\textbf{a},\textbf{b}) show a partial solution after processing nodes \{A, J\} and \{A, J, C, E, I, F\}, respectively. (\textbf{c})~shows the final solution. The last two nodes linked into the solution in the given stage of the construction process are shown in blue.}
\label{fig:ireann_example}
\end{figure}

The NN algorithm can be extended so that in each step of the tour construction process multiple choices will be considered for selecting the current working city, $s$. %
In particular, one city out of the set of not yet fully connected cities is chosen to be linked within the developed solution using the nearest neighbor heuristic. 
By the term \emph{fully connected node}, we mean the node that already has two edges---the incoming and outgoing one---assigned.
This way, multiple subtours can be developed simultaneously, which are gradually connected together resulting in the final single tour. %
The order in which cities are processed by such an \textbf{extended nearest neighbor} constructive procedure (ENN) determines the tour generated. %
When the cities are presented to (ENN) procedure in the right order, an optimal solution can be produced. This is illustrated in Figure~\ref{fig:ireann_example} where the optimal solution is produced by processing cities in the order \texttt{AJCEIFBHGD}.
IREANN is an ordinary evolutionary algorithm using a fixed-length linear representation that exploits the idea of the ENN procedure in the following way. %
Individuals evolved in the population do not represent tours directly as particular sequences of cities. %
Instead, each individual represents an input sequence to the ENN procedure called \emph{priority list} that determines the order or priorities with which the cities will be processed by the ENN procedure. %
An important aspect related to this indirect representation is that a single tour can be represented by many different priority lists. This~means there are multiple attractors to which the evolutionary algorithm can converge. On~the other hand, some tours, including the optimal one, might get unreachable with the indirect representation. Nevertheless,~any solution reachable by the standard NN procedure is reachable with the ENN procedure as well.


\textbf{IREANN algorithm}. The~pseudo-code of IREANN is shown in Algorithm~\ref{alg:ea}. It starts with a random initialization of the population of candidate priority lists (line~\ref{jies1}). Then, each individual is evaluated (line~\ref{jies2}), i.e., a tour is constructed using the ENN procedure applied to its priority list (described in Algorithm~\ref{alg:enn}) and its length is used as the individual's fitness value. After the population has been evaluated, the counter of fitness function evaluations is set to the population size (line~\ref{jies3}). Then, the algorithm runs until the number of calculated fitness evaluations reaches the maximum number of fitness evaluations (lines~\ref{jies4}--\ref{jies16}). The~first parental individual is selected in each iteration using a tournament selection (line~\ref{jies5}) and  a new individual is created using either crossover (lines~\ref{jies7}--\ref{jies10}) or mutation (lines~\ref{jies12}--\ref{jies13}) operator.
When the crossover operator is chosen with the crossover rate $P_C$, the second parental individual is selected (line~\ref{jies3}) and the offspring produced undergoes the mutation operation (lines~\ref{jies9}--\ref{jies10}). The~newly created individual is evaluated and the worst individual in the population is replaced by it (line~\ref{jies16}). %

\vspace{12pt}
\LinesNumbered
\SetKwFor{Times}{}{times do}{end}
\DontPrintSemicolon
\begin{algorithm}[H]
\small
\SetKwData{population}{$population$}
\SetKwData{parentA}{$par1$}
\SetKwData{parentB}{$par2$}
\SetKwData{offspring}{$offspring$}
\SetKwData{evaluations}{$evaluations$}
\SetKwData{goal}{$goal$}
\SetKwData{POPSIZE}{$POPSIZE$}
\SetKwData{MAXEVALS}{$MAXEVALS$}
\SetKwData{PC}{$P_C$}
\SetKwData{PM}{$P_M$}
\SetKwFunction{init}{init}
\SetKwFunction{evaluate}{evaluate}
\SetKwFunction{getBest}{getBest}
\SetKwFunction{select}{select}
\SetKwFunction{crossover}{crossover}
\SetKwFunction{mutate}{mutate}
\SetKwFunction{replace}{replace}
\SetKwFunction{rand}{rand}
\KwIn{\POPSIZE\hspace{1em}-- population size}
\myKwIn{\MAXEVALS\hspace{1em}-- maximum number of fitness function evaluations}
\myKwIn{\PC\hspace{1em}-- crossover rate}
\myKwIn{\PM\hspace{1em}-- mutation rate}
\KwOut{$tour$ -- the shortest tour found}
\vspace{-0.5em}
\nonl\hrulefill\\
\init{\population}\;\nllabel{jies1}
\evaluate{\population}\;\nllabel{jies2}
\evaluations $\leftarrow$ \POPSIZE\;\nllabel{jies3}
  \While{\evaluations $<$ \MAXEVALS} {\nllabel{jies4}
    \parentA $\leftarrow$ \select{\population}\;\nllabel{jies5}
    \If{\rand $<$ \PC} {\nllabel{jies6}
      \parentB $\leftarrow$ \select{\population}\;\nllabel{jies7}
      \offspring $\leftarrow$ \crossover{\parentA, \parentB}\;\nllabel{jies8}
      \If{\rand{} $<$ \PM} {\nllabel{jies9}
        \mutate{\offspring}\;\nllabel{jies10}
      }
    }
    \Else {\nllabel{jies11}
      \offspring $\leftarrow$ \parentA\;\nllabel{jies12}
      \mutate{\offspring}\;\nllabel{jies13}
    }
    \evaluate{\offspring}\;\nllabel{jies14}
    \replace{\population, \offspring}\;\nllabel{jies15}
    \evaluations $\leftarrow$ \evaluations $+ 1$\;\nllabel{jies16}
  }
  $tour \leftarrow$ \getBest{\population}\;\nllabel{jies17}
  \KwRet{$tour$} \;\nllabel{jies18}
\caption{Evolutionary algorithm for the optimal tour generation problem.}
\label{alg:ea}
\end{algorithm}
\vspace{12pt}

In the end, the tour generated by the best-fit individual in the population is returned (line~\ref{jies18}). %
Note that the ultimate output of the algorithm, when used to solve the GTSPC, is the candidate goal to which the robot is going to move in the next step of the environment exploration. Thus, just the first goal in the tour of the best fit individual will be returned in the end.

\textbf{IREANN evolutionary operators}. The~evolutionary algorithm uses the \emph{order-based crossover} defined in \cite{syswerda91} and a simple \emph{point mutation}.
The crossover works so that a set of goals randomly chosen from the priority list of the first parent are copied to the offspring into the same positions as they appear in the first parent.
Remaining positions are filled in with goals taken in the same order as they appear in the priority list of the second parent.

\begin{figure}[H] 
\begin{center}
\includegraphics[width=0.35\columnwidth]{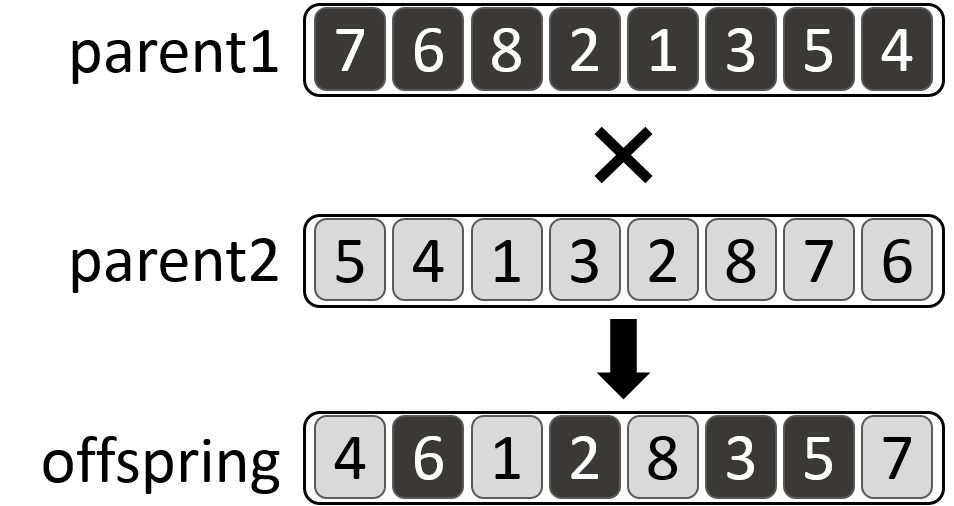}
\vspace{-1em}
\end{center}
\caption{Order-based crossover operator.}
\label{fig:crossover}
\end{figure}

The crossover is illustrated in Figure~\ref{fig:crossover}, where two priority lists $\mathrm{parent1}=[7, 6, 8, 2, 1, 3, 5, 4]$ and $\mathrm{parent2}=[5, 4, 1, 3, 2, 8, 7, 6]$ are crossed over.
First, a group of goals $\{6, 2, 3, 5\}$ is inherited from $\mathrm{parent1}$.
Then, the remaining positions of the offspring are filled in with goals $\{4, 1, 8, 7\}$ while respecting their relative order in $\mathrm{parent2}$.

IREANN uses a simple point mutation operator that randomly chooses one city in the priority list and moves it to an arbitrary position.

\subsection{IREANN Adaptation for GTSPC}
\label{sec:ireann-gtspc}

This section describes modifications related to adaptation of the IREANN algorithm to the GTSPC, namely the representation and the tour construction procedure making use of the ENN procedure and taking into account the constraint that all frontiers are fully covered.

\textbf{Representation}.
A candidate solution is a tour through a subset of candidate goals $\mathbf{G'}\subseteq\mathbf{G}$ such that the union of visibility regions of goals $g\in\mathbf{G'}$ covers all frontier cells. Generally, there might be multiple subsets of goals that produce a feasible solution. Thus, the representation should cover all the possibilities. Therefore, the representation used in this work is a \emph{priority list}, i.e. the permutation, overall candidate goals in the original set $\mathbf{G}$. %
Note that, even though the priority list contains all candidate goals, the resulting tour can be composed of just a subset of these goals, as described in the following paragraphs.

\textbf{Tour construction procedure}. 
For the sake of efficiency, the set of frontiers $\mathcal{F}$ is split into two sets---a set $\mathcal{F_N}$ of $K$ frontiers nearest to the current position of the robot and a set $\mathcal{F_D}$ of remaining distant frontiers---which are treated differently.
For each frontier in the set $\mathcal{F_D}$, a tour component $C$ is constructed using the standard nearest neighbor algorithm such that $C$ consists of a minimal set of goals completely covering the given frontier. 
The set of disjoint tour components, $\mathcal{C}$, constitutes a so-called \emph{embryo}, which is created once before the evolutionary optimization starts. The~tour components are immutable, only their connection within the whole tour is subject to further optimization. 

The embryo plus individual goals belonging to frontiers from $\mathcal{F_N}$ are passed as the input to the construction of the whole tour using the ENN procedure described in the next paragraph. 
Using the embryo leads to the search space reduction and possibly increased efficiency of the evolutionary~algorithm.

The idea behind this two-step construction strategy is that the optimal configuration of goals (i.e., their selection and interconnection) covering the frontiers close to the current robot position is crucial for the selection of the proper goal to move to in the next iteration of the exploration process. On the other hand, sub-optimally connected goals belonging to frontiers far from the starting robot position have a small impact on the next goal decision-making.

\textbf{ENN procedure}.
The ENN procedure takes a priority list $\mathrm{P}$ and the initial set of tour components $\mathcal{C}$ as input and produces a complete tour through the goals so that all frontier cells are visible from (or covered by) at least one of the used goals. The~procedure starts with the initialization of the set of uncovered frontier cells with the set of all frontier cells of uncovered frontiers (line~\ref{enn1}). The~number of available connections of each goal is set to 2 (line~\ref{enn2}). The~value of $availConn[g_i]$ indicates whether the goal $g_i$ is already fully connected in the generated tour (i.e., $availConn[g_i]=0$) or it is partly connected with one link and the other link is still available (i.e., $availConn[g_i]=1$) or it is not used in the generated tour yet (i.e., $availConn[g_i]=2$). Then, the algorithm iterates through lines~\ref{enn4}--\ref{enn31} until a single tour has been constructed that covers all frontiers' cells. In each iteration $i$, the goal to be processed $g$ is taken from the priority list $\mathrm{P}$ (line~\ref{enn5}). If $g$ is not yet used in the partial solution and the set of frontier cells visible from $g$ does not contain any of the uncovered frontier cells, then it is skipped as it does not contribute to the overall coverage of the solution (lines~\ref{enn7}--\ref{enn8}). Otherwise, $g$ is used to extend the current partial solution in three possible ways:
\begin{enumerate}[leftmargin=*,labelsep=5mm]
  \item The goal $g$ is not connected in the solution yet. Its nearest available neighbor goal $g_{nn}$ is found, the two goals $g$ and $g_{nn}$ are connected, the set of uncovered frontier cells is updated accordingly and the available connections of $g$ and $g_{nn}$ are decremented (lines~\ref{enn9}--\ref{enn14}).
  \item The goal $g$ is already linked to one other goal. Its nearest available neighbor goal $g_{nn}$ is found, the two goals $g$ and $g_{nn}$ are connected and the available connections of $g$ and $g_{nn}$ are updated (lines~\ref{enn15}--\ref{enn19}). The~set of uncovered frontier cells remains unchanged.
  \item The goal $g$ is already fully connected, i.e., it is linked to two other goals in one tour component $C^g$.
  First, the nearest available neighbors $start_{nn}$ and $end_{nn}$ of the component's boundary goals are found. Then, the shorter link of the two possible links ($start$,~$start_{nn}$) and ($end$,~$end_{nn}$) is added to the solution and the available connections of newly connected goals are updated (lines~\ref{enn20}--\ref{enn31}).
\end{enumerate}


Note that the nearest available neighbor is such a goal that is either already connected within the solution and has at least one connection available or is not used in the solution yet and can reduce the number of uncovered frontier cells at least by one.

\begin{figure}[H]
\begin{center}
\subfloat[][]{\quad\includegraphics[width=0.28\columnwidth]{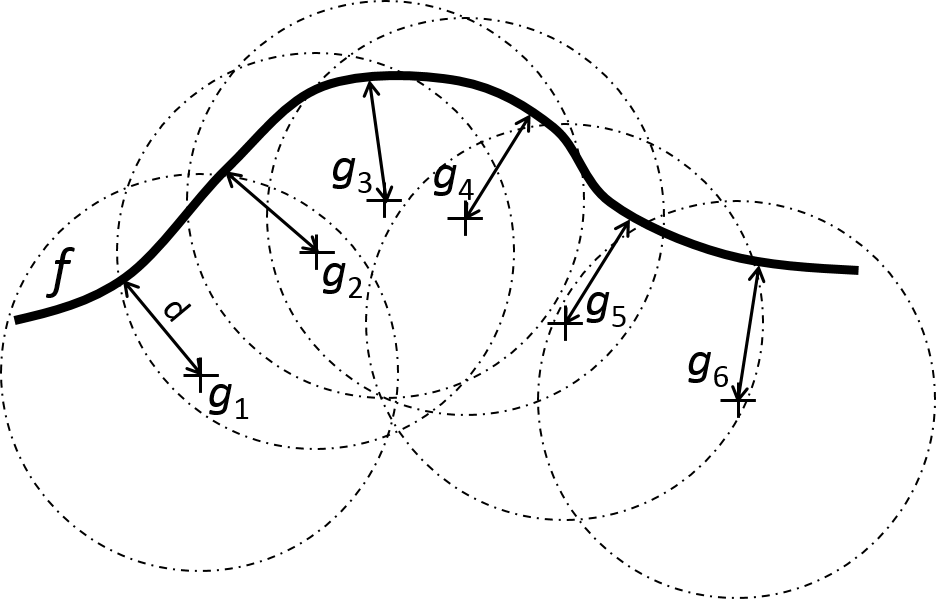}\quad\label{fig:enn_exampleA}}
\subfloat[][]{\quad\includegraphics[width=0.28\columnwidth]{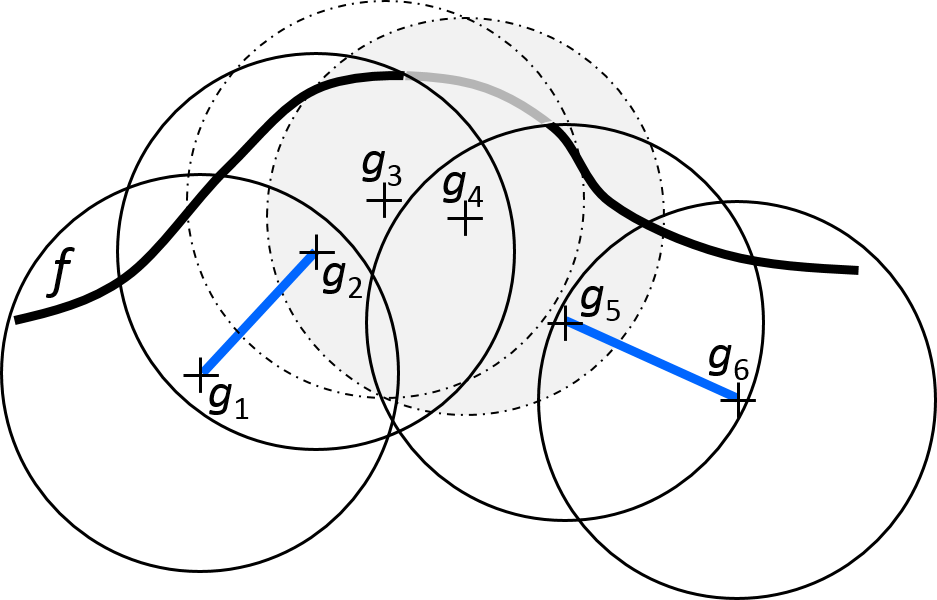}\quad\label{fig:enn_exampleB}}
\subfloat[][]{\quad\includegraphics[width=0.28\columnwidth]{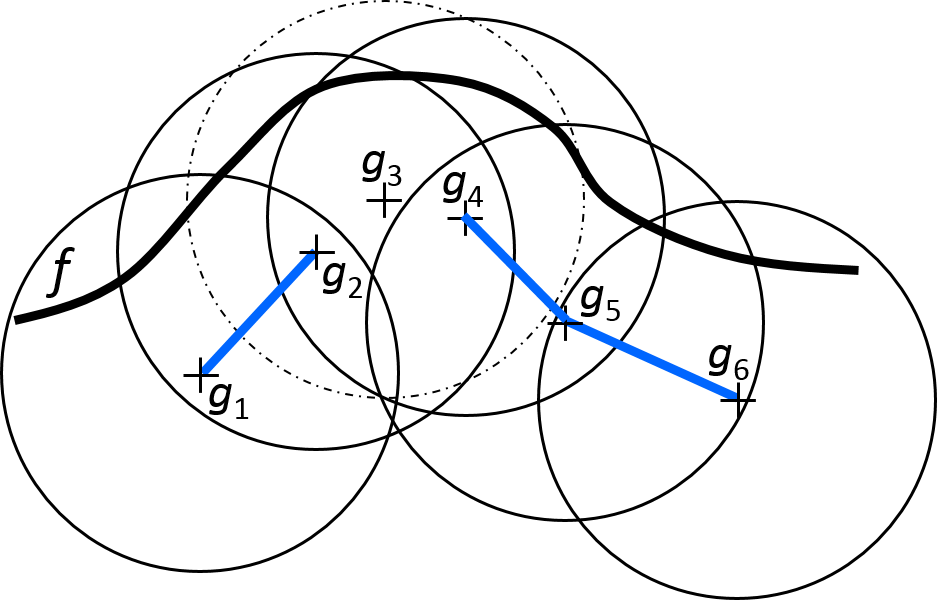}\quad\label{fig:enn_exampleC}}

\subfloat[][]{\quad\includegraphics[width=0.3\columnwidth]{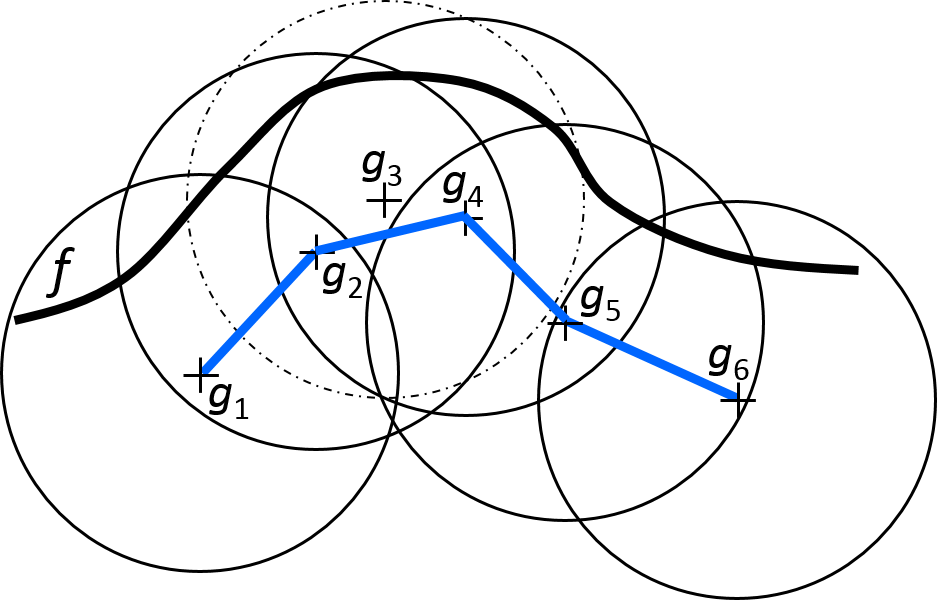}\quad\label{fig:enn_exampleD}}
\subfloat[][]{\quad\includegraphics[width=0.3\columnwidth]{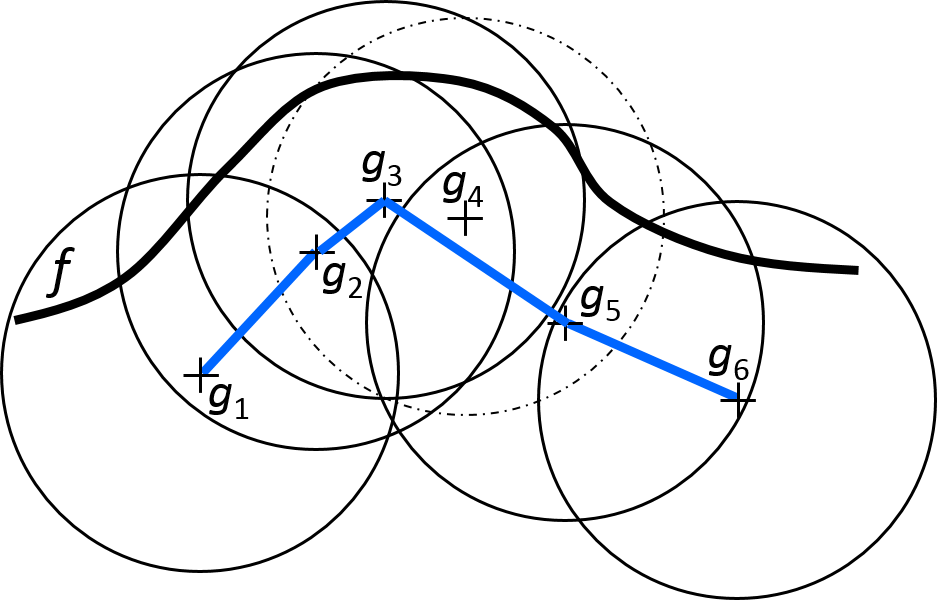}\quad\label{fig:enn_exampleE}}
\end{center}
\vspace{-1em}
 \caption{An example with a single frontier, six candidate goals equally distant from the frontier and the priority list $\mathrm{P}=\left[g_1,g_6,g_4,g_2,g_3,g_5\right]$. (\textbf{a}) the scenario; (\textbf{b},\textbf{c}) partial coverages of frontier $f$ after applying first two and three steps of the ENN 
procedure to goals $g_1$, $g_6$ and $g_4$; (\textbf{d}) the complete coverage of frontier $f$ induced by processing of the goals in the order $g_1$, $g_6$, $g_4$ and $g_2$; (\textbf{e}) an alternative coverage of frontier induced by processing of the goals in the order $g_1$, $g_2$, $g_6$ and $g_5$.}
\label{fig:enn_example}
\end{figure}

Figure~\ref{fig:enn_example} illustrates the three possible ways the partial solution can be extended.
It shows a single frontier and six candidate goals sampled for this frontier and assumes the priority list $\mathrm{P}=[g_1,g_6,g_4,g_2,g_3,g_5]$ is used to construct a tour covering this frontier.
A circle around each goal $g_i$ indicates a visibility range of the goal and the portion of a frontier covered by the circle contains the set of frontier cells covered by the goal $\mathbf{R}(g)$.
Figure~\ref{fig:enn_example}b depicts the partial solution obtained after the goals $g_1$ and $g_6$ have been processed by the ENN procedure.
Both were added to the solution by applying lines~\ref{enn9}--\ref{enn14} of Algorithm~\ref{alg:enn}.
At that point, goal $g_4$ is going to be processed by ENN procedure.
It is not yet used in the partial solution and it can still contribute to the overall coverage of the solution if added to it, so the goal will be processed using lines~\ref{enn9}--\ref{enn14} as well.
Assuming that $distance(g_4, g_5)$ is less than $distance(g_4, g_2),$ goal $g_5$ is the nearest available neighbor of $g_4$, resulting in partial solution in Figure~\ref{fig:enn_example}c.
The goal $g_3$ was not available since it does not contribute to the overall coverage of the constructed solution.
\vspace{12pt}

\begin{algorithm}[H]
 \small
\scriptsize
\SetKwData{P}{$\mathrm{P}$}
\SetKwData{Pi}{$\mathrm{P}[i]$}
\SetKwData{U}{$\mathbf{U}$}
\SetKwData{i}{$i$}
\SetKwData{D}{$\mathrm{D}$}
\SetKwData{G}{$\mathbf{G}$}
\SetKwData{F}{$\mathcal{F_U}$}
\SetKwData{C}{$\mathcal{C}$}
\SetKwData{g}{$g$}
\SetKwData{tour}{$tour$}
\SetKwData{r}{$r$}
\SetKwData{startgoal}{$start$}
\SetKwData{rightgoal}{$end$}
\SetKwData{gnn}{$g_{nn}$}
\SetKwData{gcov}{$\mathbf{R}(g)$}
\SetKwData{availConn}{$availConn$}
\SetKwData{neighborStart}{$start_{nn}$}
\SetKwData{neighborEnd}{$end_{nn}$}
\SetKwData{Cg}{$C^g$}
\SetKwFunction{init}{init}
\SetKwFunction{update}{update}
\SetKwFunction{getComponent}{getComponent}
\SetKwFunction{getNearestNeighbor}{getNearestN}
\SetKwFunction{getBoundary}{getBoundary}
\SetKwFunction{connect}{connect}
\SetKwFunction{decrement}{decrement}
\SetKwFunction{oneopt}{oneopt}
\SetKwFunction{twoopt}{twoopt}
\SetKwFunction{getFirstGoal}{getFirstGoal}
\KwIn{$\G=\{g_1,\dots, g_n\}$\hspace{1em}-- set of $n$ goals}
\myKwIn{\C\hspace{1em}-- set of initial tour components, i.e. the embryo}
\myKwIn{\F\hspace{1em}-- set of uncovered frontiers}
\myKwIn{\D\hspace{1em}-- $n\times n$ distance matrix}
\myKwIn{\P\hspace{1em}-- priority list}
\myKwIn{\r\hspace{1em}-- current robot position}
\KwOut{\tour -- tour through a subset of candidate goals}
\vspace{-0.5em}
\nonl\hrulefill\\
\U $\leftarrow$ \F\;\nllabel{enn1}
\init{\availConn}\;\nllabel{enn2}
\i $\leftarrow$ 0\;\nllabel{enn3}
  \While{\U $\neq \emptyset$} {\nllabel{enn4}
    \g $\leftarrow$ \Pi\;\nllabel{enn5}
    \i$\leftarrow \i + 1$ \;\nllabel{enn6}
    \If{$\availConn[\g] = 2$ {\bf and} $\gcov \cap \U = \emptyset$} {\nllabel{enn7}
      continue \;\nllabel{enn8}
    }
    \If{$\availConn[\g] = 2$}{\nllabel{enn9}
      \gnn $\leftarrow$ \getNearestNeighbor{\g}\;\nllabel{enn10}
      $\U \leftarrow \U \setminus (\gcov \cap \U)$ \;\nllabel{enn11}
      \connect{\g, \gnn}\;\nllabel{enn12}
      \decrement{$\availConn[\g]$}\;\nllabel{enn13}
      \decrement{$\availConn[\gnn]$}\;\nllabel{enn14}
    }
    \ElseIf{$\availConn[\g] = 1$}{\nllabel{enn15}
      \gnn $\leftarrow$ \getNearestNeighbor{\g}\;\nllabel{enn16}
      \connect{\g, \gnn}\;\nllabel{enn17}
      \decrement{$\availConn[\g]$}\;\nllabel{enn18}
      \decrement{$\availConn[\gnn]$}\;\nllabel{enn19}
    }
    \Else {\nllabel{enn20}
      \Cg $\leftarrow$ \getComponent{\g}\;\nllabel{enn21}
      \{\startgoal, \rightgoal\} $\leftarrow$ \getBoundary{\Cg}\;\nllabel{enn22}
      \neighborStart $\leftarrow$ \getNearestNeighbor{\startgoal}\;\nllabel{enn23}
      \neighborEnd $\leftarrow$ \getNearestNeighbor{\rightgoal}\;\nllabel{enn24}
      \If{$\D[\startgoal, \neighborStart] < \D[\rightgoal, \neighborEnd]$}{\nllabel{enn25}
        \connect{\startgoal, \neighborStart}\;\nllabel{enn26}
        \decrement{$\availConn[\startgoal]$}\;\nllabel{enn27}
        \decrement{$\availConn[\neighborStart]$}\;\nllabel{enn28}
      }
      \Else{
        \connect{\rightgoal, \neighborEnd}\;\nllabel{enn29}
        \decrement{$\availConn[\rightgoal]$}\;\nllabel{enn30}
        \decrement{$\availConn[\neighborEnd]$} \;\nllabel{enn31}
      }
    }
  }
  \oneopt{\tour} \;\nllabel{enn32}
  \twoopt{\tour} \;\nllabel{enn33}
  \KwRet{\tour} \;\nllabel{enn34}

\caption{Extended nearest neighbor constructive procedure for the feasible tour generation problem.}
\label{alg:enn}
\end{algorithm}

\vspace{12pt}
The next goal in the priority list is $g_2$, which is already partly connected in the solution.
\mbox{However,~it has one connection} available still.
Thus, its nearest available neighbor is found, i.e., goal $g_4$, and the two are connected resulting in the solution in Figure~\ref{fig:enn_example}d.
Theseactions correspond to lines~\ref{enn15}--\ref{enn19}.

The next goal in the priority list is $g_3$.
However, this goal is not in the solution yet, and it would not contribute to the overall coverage of the constructed solution if it were added to.
Thus, it is skipped.

The~remaining goal, $g_5$, is already fully connected in the solution and it will be processed according to lines~\ref{enn20}--\ref{enn31}.
As it cannot be connected itself to any other goal, the nearest neighbors of the boundary goals $g_1$ and $g_6$ of the component $C^{g_5}=[g_1,g_2,g_4,g_5,g_6]$ will be found.
Then, the shorter link leading from the boundary goals will be added to the solution (not shown in Figure~\ref{fig:enn_example}).

Figure~\ref{fig:enn_example}e shows another way the frontier $f$ can be covered if the priority list $\mathrm{P}=[g_1,g_6,g_4,g_2,g_3,g_5]$ was used. This time, the goal $g_4$ would be omitted.

Once a single tour covering all frontier cells is constructed, the tour is further refined by one-opt and two-opt heuristics (lines~\ref{enn32}--\ref{enn33}). These two refinement heuristics operate on a fixed set of goals and just their order can be modified.

\section{Experimental Evaluation}
\label{sec:experiments}
\vspace{-6pt}
\subsection{Simulations}

Performance of the proposed evolutionary algorithm as a goal selection strategy in the exploration framework was statistically evaluated and compared with three state-of-the-art approaches.
The~first one is Yamauchi's greedy approach (\greedy)~\cite{Yamauchi97}, probably the most popular and used strategy nowadays---see, e.g., \cite{Campos2017,Quin2017,Zheng2018DesignOA}.
The second one is our previous strategy~\cite{Kulich2011}, which minimizes the cost over several steps.
As~the number of steps corresponds to the number of candidates, we call the strategy Full Horizon Planning (\tsp).
To the best of our knowledge, \tsp{} is currently one of the best strategies which do not use background knowledge.
For example, although the recently published approach~\cite{Osswald2016} benefits from a priori known topological map of the environment, it produces similar results to \tsp{} in similar environments in comparison to \greedy.
The third method is an information-based strategy~\cite{Umari2017} that determines the cost of a candidate as a weighted sum of the information gained from visiting the goal and the Euclidean distance to the candidate. A hysteresis gain is added to prefer goals in robot's vicinity. 
This method is further referred to as \umari{}.

The methods differ in the evaluation of goal candidates as well as in the generation of these.
While \greedy{} selects the nearest cell among all frontiers cells, \tsp{} clusters frontier cells using k-means, which is also used in our implementation of \umari{}. 
The proposed approach (\ga) employs a two-stage sampling process as described in Section~\ref{sec:framework}.

The comparison follows Level-0 of the methodology presented in~\cite{Faigl2015}, which suggests studying theoretical behavior of methods without the influence of sensor noise, localization imprecision, and~inaccuracies of motion control.
Therefore, we employ our robotic simulator, in which the complete exploration framework with the above-mentioned strategies integrated was implemented in C++ except the evolutionary algorithm itself, which was done in Java.

Three maps from Motion Planning Maps Dataset~\cite{mapdataset} scaled to $20$~m$~\times~20$~m representing various types of environments were chosen for comparison.
The first one (\emp) is a map without obstacles giving a robot high flexibility in motion.
The \potholes{} map represents an unstructured space with about 20 small obstacles.
Finally, \jari{} is a map of an administrative building with many rooms and corridors between~them.

A sensor with $360^\circ$ field of view with various visibility ranges $\rho\in\{1.5, 2.0, 3.0, 5.0, 10.0\}$ (in~meters) was
used, while an occupancy grid with cell size $0.05\times 0.05~m$ was chosen to represent the working environment.
Robot size was $0.1$~m, the inflation radius was, $d=0.25$~m, the distance for the uniform sampling was set to $0.25~m$, and every fourth cell was considered as a goal candidate (constant $k$ in Algorithm~\ref{sc:framework}).

The evolutionary algorithm was run with the following configuration:  population size = 200, maximal number of fitness evaluations = 3000, crossover rate = 80\%, mutation rate = 25\%, number of nearest frontiers $K$ = 5, and tournament size = 3.
For the \umari{ strategy}, the settings recommended by the authors is used: $\lambda=3$, $h_{gain} = 2$, $h_{rad} =$ sensor range (see~\cite{Umari2017} and \url{https://github.com/hasauino/rrt_exploration}).

Fifty trials were run for each combination $\left<\text{map, range, method}\right>$, which gives 2250 trials in total.
The~obtained results for \greedy{}, \tsp{} and \ga{} are statistically summarized in Table~\ref{table1}, where
{\em avg} stands for the time needed to explore the whole environment (exploration time $T_{exp}$) averaged over all 50~runs, {\em min} and {\em max} are minimal and maximal $T_{exp}$ over these runs and {\em stdev} stands for the standard deviation of $T_{exp}$.
{$R_{gr}$} expresses a ratio of the average $T_{exp}$ obtained with the proposed method to the average $T_{exp}$ obtained with the \greedy{} one. Similarly, {$R_{FHP}$} is a ratio of average $T_{exp}$ values of \ga{} and \tsp{}. A value of less than 100\% indicates the proposed approach is better than the respective compared one, and vice versa. Differences between the average $T_{exp}$ values of the compared algorithms were evaluated using a two-sample $t$-test with the significance level $\alpha = 0.01$. The~null hypothesis being the two data vectors are from populations with equal means. Results of the statistical tests are presented in the last column of the table. A sign `+' means the average value obtained with the proposed algorithm is significantly better than the one of the compared algorithm. A~sign `-' indicates the opposite case. A~situation when the two compared means are statistically indifferent is indicated by the `=' sign.

\begin{figure}[H]
\begin{center}
\subfloat[][]{\includegraphics[width=0.45\columnwidth]{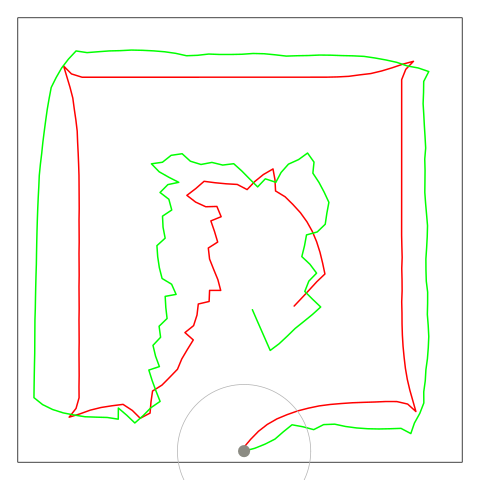}\label{fig:empty}}
\subfloat[][]{\includegraphics[width=0.45\columnwidth]{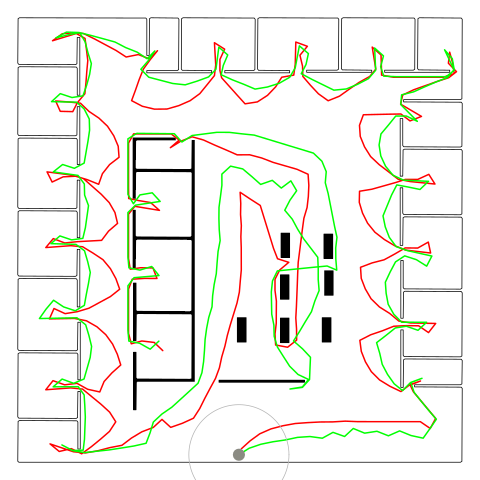}\label{fig:jari}}
\vspace{-1.7em}
\end{center}
\caption{The best results found {by} \ga{} (red) and \tsp{} (green) on 
(\textbf{a}) \emp{} with $\rho=3.0$~m  and~(\textbf{b})~\jari{} with $\rho=2.0$~m.}
\label{fig:results}
\end{figure}

It can be seen that \tsp{} and \ga{} significantly outperform the \greedy{} approach for the \emp{} map as they benefit from longer planning horizon allowing them to explore the space systematically.
Moreover, \ga{} provides better results than \tsp{}, especially for $\rho=3.0$, where the difference is more than 22\%.
This~is because \ga{} directs a robot to a distance $\rho$ from obstacles contrary to \tsp{}, where k-means generates goals to be visited closer to obstacles---see Figure~\ref{fig:results}a.

\newpage

\newpage
\paperwidth=\pdfpageheight
\paperheight=\pdfpagewidth
\pdfpageheight=\paperheight
\pdfpagewidth=\paperwidth
\newgeometry{layoutwidth=297mm,layoutheight=210 mm, left=2.7cm,right=4.1cm,top=1.8cm,bottom=1.5cm, includehead,includefoot}
\fancyheadoffset[LO,RE]{0cm}
\fancyheadoffset[RO,LE]{0cm}

\begin{table}
\centering
\vspace{-7em}
\caption{Comparison of the exploration strategies.} \label{table1} 
\scalebox{0.9}[0.9]{\begin{tabular}{ccp{0.01em}rrrrp{0.01em}rrrrp{0.01em}rrrrp{0.01em}rrc}
\toprule
\textbf{Map} & \boldmath{$\rho$} & & \multicolumn{4}{c}{\textbf{Greedy}} & & \multicolumn{4}{c}{\textbf{FHP-Based}} & & \multicolumn{4}{c}{\textbf{EA-Based}} & & \cc{\boldmath{$R_{gr}$}} & \cc{\boldmath{$R_{\textbf{FHP}}$}} & \textbf{+/}\boldmath{$-$}  \\
&~~ & & \cc{\textbf{avg}} & \cc{\textbf{min}} & \cc{\textbf{max}} &  \cc{\textbf{stdev}} & & \cc{\textbf{avg}} & \cc{\textbf{min}} & \cc{\textbf{max}} & \cc{\textbf{stdev}} & & \cc{\textbf{avg}} & \cc{\textbf{min}} & \cc{\textbf{max}} & \cc{\textbf{stdev}} &  & \textbf{\%} & \textbf{\%}\\
\midrule
\multirow{5}{*}{empty}
  & 1.5 & & 2419.70 & 2171.00 & 2701.00 & 111.43 && 2241.90 & 2041.00 & 2436.00 & 89.45 && 2118.60 & 1911.00 & 2226.00 & 68.67 & & 87.56 & 94.50 & {\cplus}/{\cplus}\\
  & 2.0 & & 1924.90 & 1741.00 & 2186.00 & 82.77 && 1679.90 & 1441.00 & 1816.00 & 86.97 && 1524.70 & 1381.00 & 1621.00 & 45.78 & & 79.21 & 90.76 & {\cplus}/{\cplus}\\
  & 3.0 & & 1330.70 & 1101.00 & 1556.00 & 91.78 && 1050.40 & 966.00 & 1196.00 & 64.12 && 918.50 & 881.00 & 966.00 & 18.77 & & 69.02 & 87.44 & {\cplus}/{\cplus}\\
  & 5.0 & & 739.90 & 621.00 & 856.00 & 65.79 && 585.20 & 571.00 & 621.00 & 9.44 && 542.20 & 516.00 & 566.00 & 10.91 & & 73.28 & 92.65 & {\cplus}/{\cplus}\\
  & 10.0 & & 428.50 & 341.00 & 471.00 & 39.70 && 305.30 & 286.00 & 326.00 & 7.83 && 294.90 & 271.00 & 361.00 & 13.07 & & 68.82 & 96.59 & {\cplus}/{\cplus}\\
\midrule
\multirow{5}{*}{potholes}
  & 1.5 & & 2604.80 & 2361.00 & 2886.00 & 125.88 && 2391.80 & 2171.00 & 2506.00 & 78.33 && 2172.50 & 2021.00 & 2311.00 & 69.24 & & 83.40 & 90.83 & {\cplus}/{\cplus}\\
  & 2.0 & & 2021.40 & 1846.00 & 2321.00 & 105.73 && 1816.40 & 1581.00 & 1931.00 & 73.48 && 1724.60 & 1606.00 & 1856.00 & 47.20 & & 85.32 & 94.95 & {\cplus}/{\cplus}\\
  & 3.0 & & 1293.60 & 1271.00 & 1461.00 & 36.65 && 1321.00 & 1141.00 & 1451.00 & 75.07 && 1259.20 & 1176.00 & 1346.00 & 42.91 & & 97.34 & 95.32 & {\cplus}/{\cplus}\\
  & 5.0 & & 1032.90 & 1011.00 & 1141.00 & 27.86 && 957.40 & 876.00 & 1056.00 & 34.70 && 911.90 & 856.00 & 951.00 & 23.20 & & 88.29 & 95.25 & {\cplus}/{\cplus}\\
  & 10.0 & & 951.30 & 741.00 & 1026.00 & 94.23 && 684.40 & 631.00 & 771.00 & 32.27 && 696.90 & 641.00 & 796.00 & 29.18 & & 73.26 & 101.83 & {\cplus}/\bf = \\
\midrule
\multirow{5}{*}{jari-huge}
  & 1.5 & & 2703.70 & 2511.00 & 2916.00 & 69.50 && 2331.40 & 1831.00 & 2391.00 & 77.58 && 2392.80 & 2331.00 & 2461.00 & 26.66 & & 88.50 & 102.63 & {\cplus}/\cminus \\
  & 2.0 & & 2022.90 & 1841.00 & 2281.00 & 135.65 && 1776.40 & 1711.00 & 1841.00 & 27.51 && 1855.70 & 1826.00 & 1896.00 & 21.34 & & 91.73 & 104.46 & {\cplus}/\cminus\\
  & 3.0 & & 1370.00 & 1256.00 & 1521.00 & 93.48 && 1194.60 & 1136.00 & 1236.00 & 36.20 && 1190.20 & 1146.00 & 1251.00 & 30.61 & & 86.88 & 99.63 & {\cplus}/\bf = \\
  & 5.0 & & 1254.70 & 1166.00 & 1331.00 & 70.15 && 1055.10 & 986.00 & 1101.00 & 28.51 && 1079.60 & 1041.00 & 1101.00 & 15.05 & & 86.04 & 102.32 & {\cplus}/\cminus\\
  & 10.0 & & 1168.50 & 1136.00 & 1201.00 & 14.85 && 952.90 & 896.00 & 1001.00 & 22.56 && 973.30 & 926.00 & 1086.00 & 39.80 & & 83.29 & 102.14 & {\cplus}/\cminus\\
\midrule
\end{tabular}}

\end{table}

\newpage
\restoregeometry
\paperwidth=\pdfpageheight
\paperheight=\pdfpagewidth
\pdfpageheight=\paperheight
\pdfpagewidth=\paperwidth
\headwidth=\textwidth



The situation is similar for \potholes{}, where \greedy{} is outperformed by the sophisticated  approaches, although the difference is not as big as for \emp.
In addition, \ga{} gives better results than \tsp{} in all cases except one.

Finally, \ga{} performs better by approx. 8--15\% than \greedy{} for \jari{}.
On the other hand, it is slightly (up to 5\%) outperformed by \tsp{} due to the same reason that it is better than \tsp{} for \emp.
Here,~directing the robot far from obstacles when going between neighboring rooms is contra-productive (see in Figure~\ref{fig:results}b).
Conversely, \ga{} is more effective in the empty area in the middle.

The~results for \umari{} are presented in Table~\ref{table2}.
The~meaning of the symbols is the same as in Table~\ref{table1} except {$R_{gr}$}, which expresses a ratio of the average $T_{exp}$ obtained with \umari{} to the average $T_{exp}$ obtained with \greedy{}. Similarly, {$R_{EA}$} is a ratio of average $T_{exp}$ values of \umari{} and \ga{}.
It can be seen that the performance of \umari{} is the worst in all cases even in comparison to \greedy{}.  

\begin{table}[H]
\begin{center}
\caption{Experimental results for \umari{}.} \label{table2}
\begin{tabular}{ccp{0.01em}rrrrp{0.01em}rr}
\toprule
\textbf{Map} & \boldmath{$\rho$} & & \multicolumn{4}{c}{\textbf{Greedy}} & &  \cc{\boldmath{$R_{gr}$}} & \cc{\boldmath{$R_{EA}$}} \\
&~~ & & \cc{\textbf{avg}} & \cc{\textbf{min}} & \cc{\textbf{max}} &  \cc{\textbf{stdev}} & & \textbf{\%} & \textbf{\%}\\
\midrule
\multirow{4}{*}{empty}
  & 1.5 & & 2766.21 & 2506.00 & 3066.00 & 132.46 & & 114.32  & 130.57\\
  & 2.0 & & 2196.61 & 1971.00 & 2431.00 & 112.40 & & 114.12  & 144.07\\
  & 3.0 & & 1596.71 & 1406.00 & 1816.00 & 101.32 & & 119.99  & 173.84\\
  & 5.0 & & 1290.15 & 1136.00 & 1446.00 & 77.96 & & 174.37  & 237.95\\
\midrule
\multirow{4}{*}{potholes}
  & 1.5 & & 2711.93 & 1966.00 & 3026.00 & 268.21 & & 104.11  & 124.83\\
  & 2.0 & & 2198.11 & 1451.00 & 2441.00 & 194.48 & & 108.74  & 127.46\\
  & 3.0 & & 1495.38 & 436.00 & 2026.00 & 402.12 & & 115.60  & 118.76\\
  & 5.0 & & 1429.27 & 471.00 & 1686.00 & 189.93 & & 138.37  & 156.73\\
\midrule
\multirow{4}{*}{jari-huge}
  & 1.5 & & 2845.89 & 2651.00 & 3061.00 & 91.46 & & 105.26  & 118.94\\
  & 2.0 & & 2447.43 & 2191.00 & 2651.00 & 102.75 & & 120.99  & 131.89\\
  & 3.0 & & 1738.65 & 1571.00 & 1866.00 & 65.00 & & 126.91  & 146.08\\
  & 5.0 & & 1596.52 & 1511.00 & 1661.00 & 41.87 & & 127.24  & 147.88\\
\midrule
\end{tabular}
\end{center}
\label{tbl:umari}
\end{table} 
\unskip
An indirect comparison can also be done with recent (deep) learning-based strategies, as the authors of these strategies present a comparison to the greedy approach.
Chen et al.~\cite{chen2018learning} show that the greedy approach outperforms imitation learning, reinforcement techniques as well as Curiosity-based Exploration~\cite{pathak18largescale} when localization error is below 3\% (which is the case we are focused on). 
On~the other hand, their strategy performs better than \greedy{} when localization error increases.  
\mbox{Zhu et al.~\cite{Zhu2018}} present the evaluation of their strategy based on Reinforcement Learning supervised Bayesian Optimization on ten office plans (similar to the \jari{} map but smaller). 
Their method is worse than greedy in four cases (by 41\% in one case), while better in the other cases (by up to 32\%). 
Table~\ref{table1} shows that \ga{}, on the other hand, performs better than \greedy on \jari{} in all cases by 8--17\%.

\subsection{Time Complexity}

\begin{figure}[H]
\begin{center}
\subfloat[][]{\includegraphics[width=0.43\columnwidth]{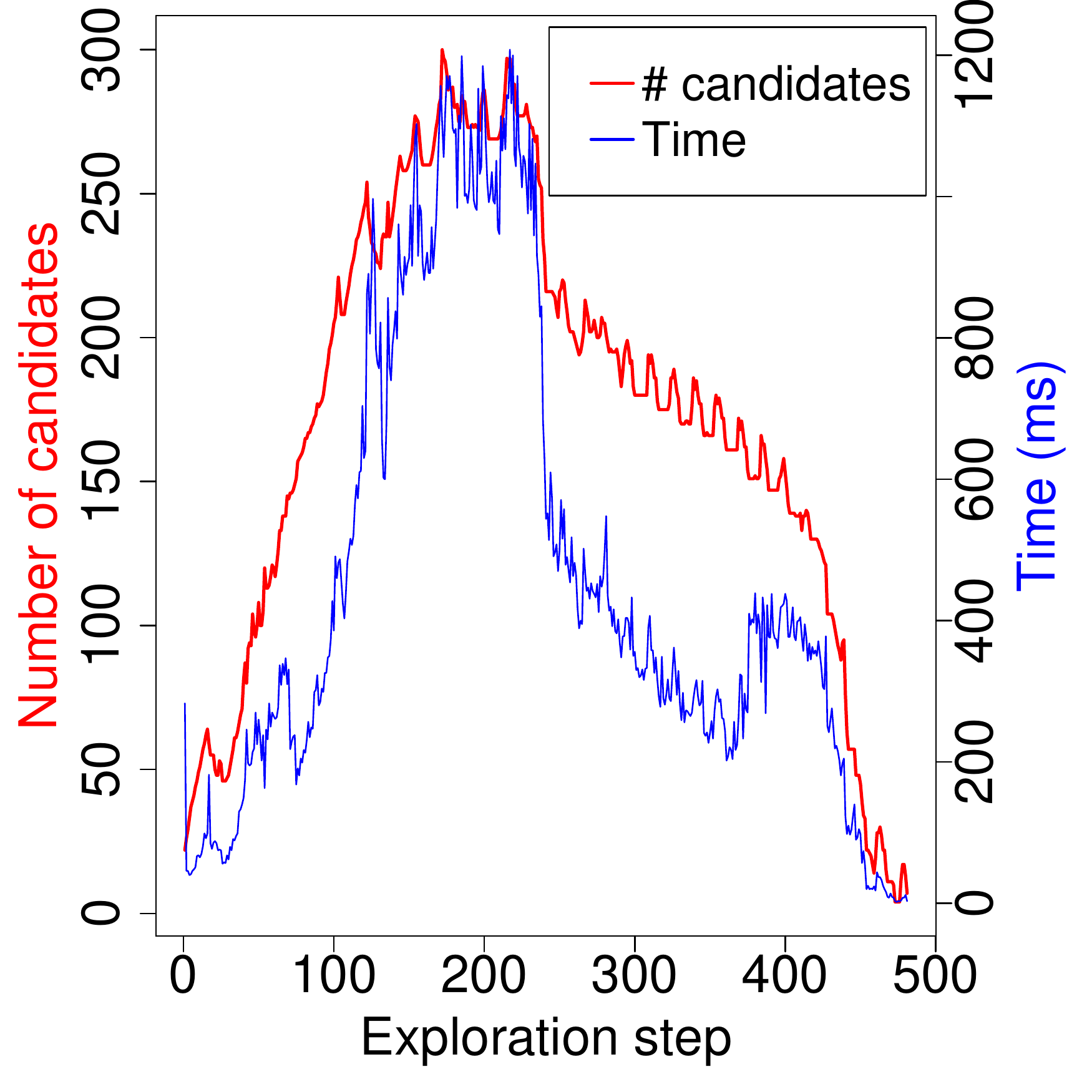}\label{fig:times-jari}}
\subfloat[][]{\includegraphics[width=0.43\columnwidth]{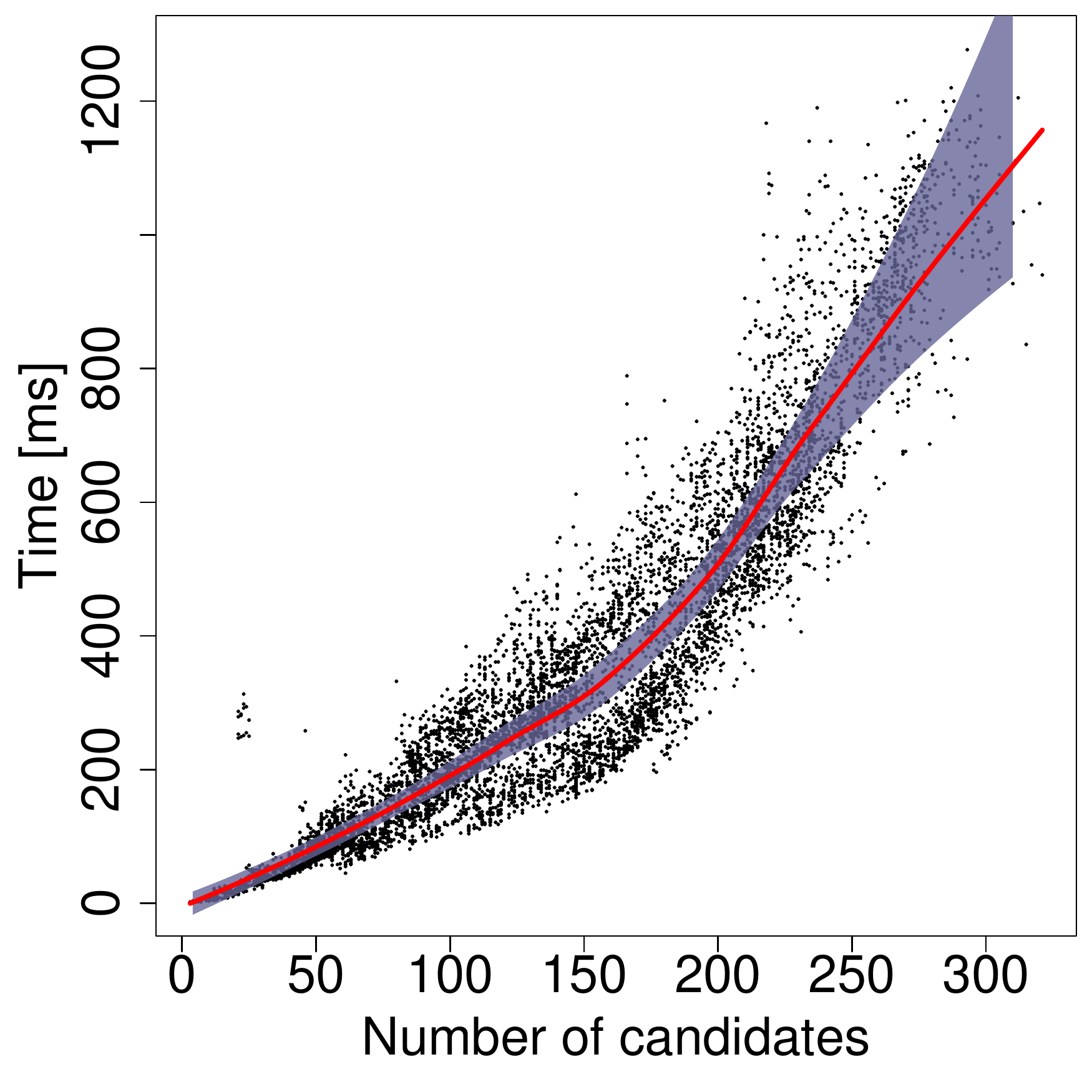}\label{fig:times-15}}
\vspace{-1.7em}
\end{center}
\caption{(\textbf{a}) evolution of computational time of EA and the number of goal candidates during a typical run on \jari{} with $\rho = 1.5$~m; (\textbf{b}) dependency of computational time of EA on a number of candidates.
The dots represent measurements from 15 runs on \jari{} with $\rho = 1.5$~m.
The red curve is a smoothed computational time averaged over the number of candidates, while the grey area represents the confidence interval computed for mean on a 95\% confidence level.
}
\label{fig:times}
\end{figure}

A set of experiments was performed to evaluate time complexity of the proposed EA algorithm on a workstation with the Intel\textsuperscript{\textregistered} Core\texttrademark{} i7-3770 CPU (Intel Corporation, Santa Clara, CA, USA) at 3.4 GHz running Sabayon Linux with the kernel 3.19.0.
Fifteen trials of full exploration were run in the most complex setup: the \jari{} map and the visibility range $\rho=1.5$~m, while the other parameters were set to the same values as in the previous case.
Figure~\ref{fig:times}a shows how the number of candidates (NoC) and computational time ($T$) of EA change during a typical trial.
It can be seen that approximately 500 runs of EA were executed during the trial and that NoC grows as the robot explores new areas at the beginning, while it starts to decrease in the middle of the process.
The curve of computational time follows the one of NoC but not so exactly as one would expect.
This~is caused by the fact that the complexity of the GTSPC problem does not only depend on NoC, but also on the number of clusters and distribution of candidates over them.
The~number of clusters varied up to 22 in this particular case.

Data from all 15 runs containing 7156 executions of EA are shown in Figure~\ref{fig:times}b in the form of dependency of $T$ on NoC together with the averaged times for particular NoCs and the confidence interval computed for the means on a 95\% confidence level. In fact,  the averages and the confidence intervals are too noisy, so local polynomial fitting is applied to make the curves smooth. 
The maximal number of candidates was 321, while computational time does not exceed 1.3~s.
This qualifies the method to be deployed in real time.

\subsection{Real Deployment}

\begin{figure}[H]
\centering 
\scalebox{0.98}[0.98]{\subfloat[][]{\includegraphics[height=3.7cm]{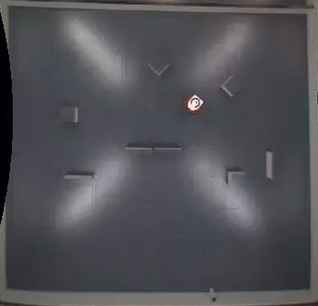}\label{fig:real-syr}}~
\hfill
\subfloat[][]{\includegraphics[height=3.7cm]{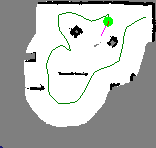}\label{fig:real1}}~
\hfill
\subfloat[][]{\includegraphics[height=3.7cm]{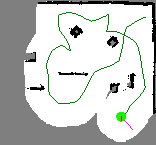}\label{fig:real2}}~
\hfill
\subfloat[][]{\includegraphics[height=3.7cm]{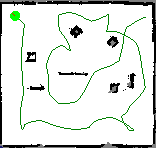}\label{fig:real3}}}
\caption{Experiment with a real robot in the SyRoTek system. (\textbf{a})~the robot during an experiment; (\textbf{b},\textbf{c})~phases of the exploration process; (\textbf{d})~the final map and the performed path.}
\label{fig:real}
\end{figure}

The whole exploration framework was deployed on a real robot in the SyRoTek system ~\cite{Kulich2013Syrotek} developed at Czech Technical University to~demonstrate the applicability of the proposed solution with real hardware.
SyRoTek is a platform for e-learning and distant experimentation in robotics and related areas consisting of thirteen robots equipped with standard robot sensors (laser range-finders, sonars, odometry, etc.).
The SyRoTek robot is called S1R and its body consists of the main chassis and an optional sensor module~\cite{Chudoba2011}. 
The robot is based on a differential drive with the maximal velocity designed to 0.35 m/s and operating time 8 h. 
The on-board computer (OBC) is the Gumstix Overo Fire module with the ARM
Cortex-A8 OMAP3530 processor unit (ARM, Cambridge, UK) operating at 600 MHz and running the Linux kernel. 
The connection with the control computer is provided by the integrated WiFi module of the Overo board.
The robots operate in the Arena of size $3.5\times 3.8$~m and are fully programmable and remotely controlled.
A HOKUYO URG-04LX laser range finder {(Hokuyo Automatic Co., Ltd., Osaka, Japan)} with a sensing view $240^\circ$ and a range limited to $0.5$~m was used as a main sensor for the experiment.
Resolution of the occupancy grids was set to 2 cm.
A Smooth Nearness Diagram (SND) algorithm~\cite{durham08} was used to control the robot motion and to avoid obstacles.
Robot position was taken from the localization system provided by SyRoTek, which is capable of continuous and errorless operation with the precision of $\pm$1~cm in robot position and $\pm3^\circ$ in robot orientation.
Figure~\ref{fig:real} shows several phases of the exploration with the EA as the goal selection strategy. 

\section{Conclusions and Future Work}
\label{sec:conclusion}

This article presents a new approach to the goal selection task solved within the robotic exploration of an unknown environment.
While state-of-the-art approaches generate a relatively small set of goal candidates from which a next goal is chosen based on some evaluation function, our approach integrates goal candidates generation with a selection of the next goal, which leads to a solution of the constrained Generalized Traveling Salesman Problem. 
The proposed approach thus allows higher flexibility in the planning of the next robot actions.     

The overall exploration framework including the proposed goal selection strategy is evaluated in a simulation environment on three different maps and compared with three state-of-the-art techniques.
The results show that the proposed method can compute results for robotic problems of a standard size in less than 1300~ms, which is sufficient for real-time usage. 
It statistically significantly outperforms the other three strategies in empty environments and areas with low density of obstacles: exploration times are by more than $30\%$ lower than for the widely used \greedy{} approach and up to $12.5\%$ lower than for \tsp{} in some cases.
On the other hand, the proposed method is worse than \tsp{} in narrow corridors by up to $4.5\%$, but still better than \greedy{} by more than $10\%$ on average. \umari{} is even worse. 
In general, the method exhibits the best overall performance.
Thus, our approach is a good choice when the type of the environment to be explored is not known in advance.

In future work, we want to focus on the generation of goal candidates.
A density of generated candidates can be controlled according to several criteria: distance to the robot, distance to obstacles, a~shape of the environment in the vicinity of a candidate, topology of the environment, experience from previous runs of GTSPC, etc.
Generation of candidates especially at places where it is interesting can influence both the quality of a GTSP solution as well as computational complexity of GTSPC.
We~also want to extend the method for multiple robots.

The~evolutionary algorithm itself can be optimized as well. In the current implementation, when evaluating a particular priority list, it is first translated to the corresponding tour, which is then locally improved by 1-opt and 2-opt heuristics.
However,~the optimized tour is only used to assess the quality of the priority list.
The final order of goals within the tour is not stored for later use.
Experience from the field of memetic algorithms suggests that it might be beneficial for the efficiency of the evolutionary process to keep the locally optimized solution instead of the original one.
Thus, we will investigate possibilities to keep the optimized tour along with the priority list in the individual and reuse it within the crossover~operator.

%

\vspace{6pt} 



\authorcontributions{M.K. developed the idea, designed the exploration framework, and performed the experiments. J.K. developed the IREANN algorithm, while L.P. supervised the work.}

\funding{This work has been supported by the European Union's Horizon 2020 research and innovation programme under Grant No. 688117 and by the European Regional Development Fund under the project Robotics for Industry 4.0 (reg. no. CZ.02.1.01/0.0/0.0/15\_003/0000470).}

\acknowledgments{Access to computing and storage facilities owned by parties and projects contributing to the National Grid Infrastructure MetaCentrum, provided under the programme \enquote{Projects of Large Infrastructure for Research, Development, and Innovations} (LM2010005), is greatly appreciated.}

\conflictsofinterest{The authors declare no conflict of interest.}

\reftitle{References}




\end{document}